\documentclass[final]{cvpr}

\usepackage{times}
\usepackage{epsfig}
\usepackage{graphicx}
\usepackage{amsmath}
\usepackage{amssymb}

\usepackage{lipsum}
\usepackage{mathtools}
\usepackage{cuted}
\usepackage{breqn}
\usepackage[pagebackref=true,breaklinks=true,colorlinks,bookmarks=false]{hyperref}

\def\cvprPaperID{11516} 
\def\confYear{CVPR 2021}

\begin{document}

\title{Multi Modal Adaptive Normalization for Audio to Video Generation}
\author{
Neeraj Kumar\\
{\tt\small neerajku@hike.in}
\and
Srishti Goel \\
{\tt\small srishtig@hike.in}
\and 
Ankur Narang\\
{\tt\small ankur@hike.in}
\and
Brejesh Lall \\
{\tt\small brejesh@ee.iitd.ac.in}
}

\maketitle

\begin{abstract}
Speech-driven facial video generation has been a complex problem due to its multi-modal aspects namely audio and video domain. The audio comprises lots of underlying features such as expression, pitch, loudness, prosody(speaking style) and facial video has lots of variability in terms of head movement, eye blinks, lip synchronization and movements of various facial action units along with temporal smoothness. Synthesizing highly expressive facial videos from the audio input and static image is still a challenging task for generative adversarial networks.  

In this paper, we propose a multi-modal adaptive normalization(MAN) based architecture to synthesize a talking person video of arbitrary length using as input: an audio signal and a single image of a person. The architecture uses the multi-modal adaptive normalization, keypoint heatmap predictor, optical flow predictor and class activation map\cite{cam} based layers to learn movements of expressive facial components and hence generates a highly expressive talking-head video of the given person. 

The multi-modal adaptive normalization uses the various features of audio and video such as Mel spectrogram, pitch, energy from audio signals and predicted keypoint heatmap/optical flow and a single image to learn the respective affine parameters to generate highly expressive video. Experimental evaluation demonstrates superior performance of the proposed method as compared to Realistic Speech-Driven Facial Animation with GANs(RSDGAN)~\cite{Alpher05}, Speech2Vid~\cite{Chung17b}, and other approaches, on multiple quantitative metrics including: SSIM (structural similarity index), PSNR (peak signal to noise ratio), CPBD (image sharpness), WER(word error rate), blinks/sec and LMD(landmark distance). Further, qualitative evaluation and Online Turing tests demonstrate the efficacy of our approach.

\end{abstract}

\section{Introduction}
Speech-driven video generation has countless applications across areas of interest including movie production, photography, e-commerce, advertisement and education. It is a complex problem that takes into account various details like the style of video, temporal consistencies, facial expressions and movements of facial action units. The information described by an image of a video generally consists of spatial and style information, such as object shape and texture. While the spatial information usually represents the key contents of the image, the style information often involves extraneous details that complicate the generation tasks. Temporal consistencies between the frames that handle natural blinks, smooth movement of facial action units and lip synchronization also play an important role in the synthesis of realistic videos.

Moderating the variability of image styles has been studied to enhance the quality of images generated from neural networks. In this regard, variants of normalization have been used to capture various information such as style, texture, shape, etc. Instance Normalization (IN)\cite{isnorm} is a representative approach which was introduced to discard instance-specific contrast information from an image during style transfer. Inspired by this, Adaptive Instance normalization\cite{ADAIN} provided a rational interpretation that IN performs a form of style normalization, showing that by simply adjusting the feature statistics—namely the mean and variance—of a generator network, one can control the style of the generated image. IN dilutes the information carried by the global statistics of feature responses while leaving their spatial configuration only, which can be undesirable depending on the task at hand and the information encoded by a feature map. To handle this, Batch-Instance Normalization(BIN)\cite{batchins} normalizes the styles adaptively to the task and selectively to individual feature maps. It learns to control how much of the style information is propagated through each channel of features leveraging a learnable gate parameter. For style transfer across the domain, UGATIT\cite{ugatit} has used adaptive instance and layer normalization\cite{LN}(LN) which adjusts the ratio of IN and LN to control the amount the style transfer from one domain to other domains. 

For style transfer tasks, a popular methodology is trying the denormalization to the learned affine transformation that is parameterized based on a separate input image (the style image). SPADE \cite{park2019SPADE}  makes this denormalization spatially sensitive. SPADE normalization boils down to "conditional batch normalization which varies on a per-pixel basis". It is implemented as a two-layer convolution neural network (compare with batch normalization, which is a simple functional layer). In the World Consistent video to video synthesis \cite{WCW}, they have used optical features and semantic maps in the normalization to learn the affine parameters to generate the realistic and temporally smoother videos. 

All the above normalization techniques have worked on capturing the styles of image and no work is done to capture the styles of audio and its mutual dependence on images in multi-modal applications through normalization. In this paper, we propose multi-modal adaptive instance normalization in the proposed architecture to generate realistic videos. We have built the architecture based on \cite{Kumar_2020_CVPR_Workshops} to show how multi-modal adaptive normalization helps in generating highly expressive videos using the audio and person's image as input.

\begin{figure}[h!]
  \includegraphics[width=\linewidth]{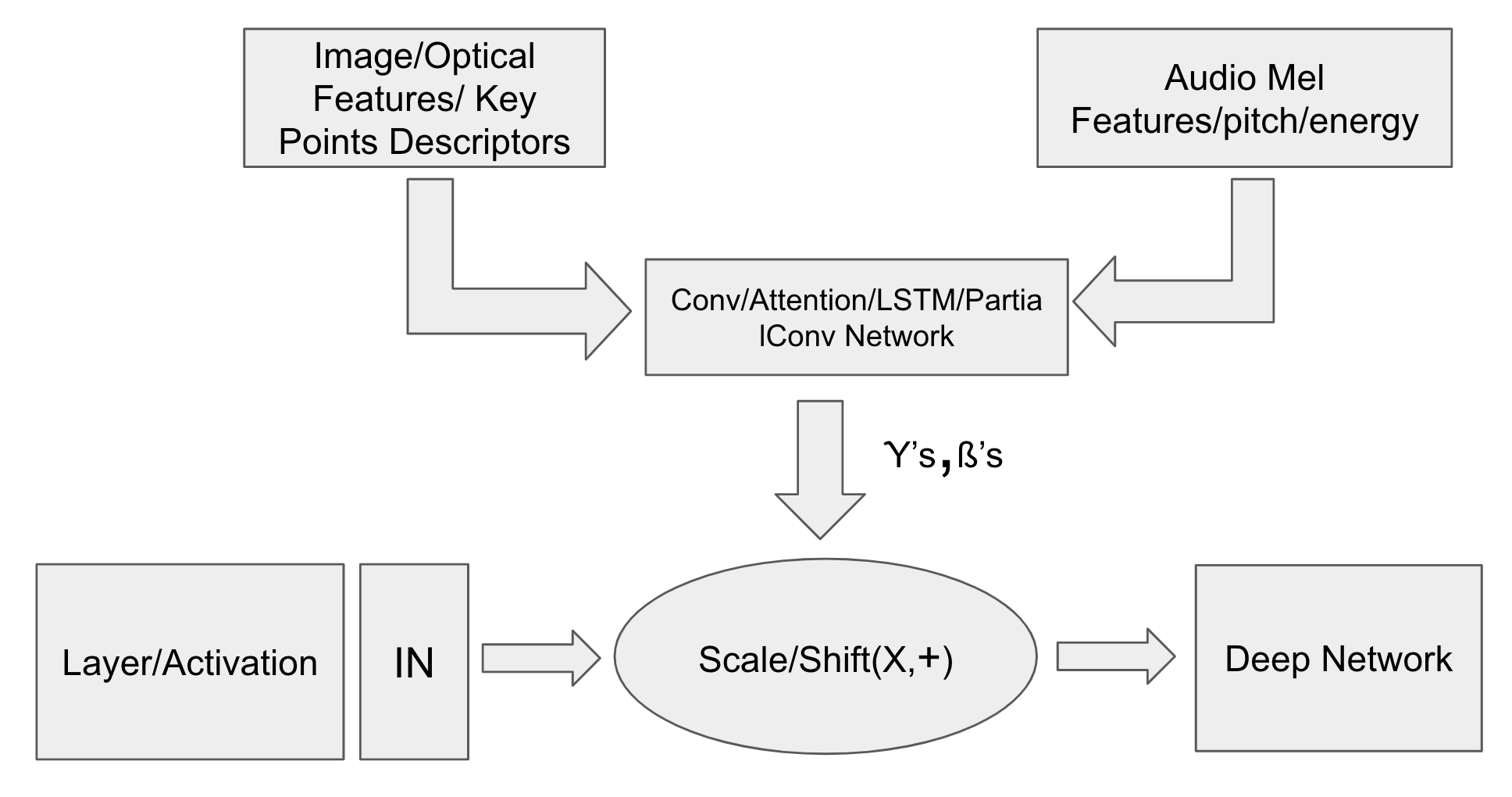}
  \caption{Higher level architecture of Multi Modal Adaptive Normalization}
  \label{fig:manres234}
\end{figure}

Our main contributions are :

\begin{itemize}
\item The affine parameters i.e, scale, $\gamma$ and a shift, $\beta$ are typically been used to learn the higher-order statistics of image features corresponding to style, texture, etc., to generate the required output as depicted in various previous works \cite{isnorm, stylegan, park2019SPADE, ugatit, WCW, batchins}. We are the first one to propose how affine parameters help to learn the higher-order statistics of multiple domains. Figure~\ref{fig:manres234} shows the higher-level architectural design of multi-modal adaptive normalization. In this paper, we have proposed the multi-modal adaptive normalization in which the affine parameters learns the higher-order information related to texture, temporal smoothness, style, lip synchronization and head movement from various features of speech namely Mel spectrogram, pitch and energy and video i.e. image frame, optical flow/keypoint heatmap to generate a realistic video from audio and a static image as an input.

\item The respective affine parameters i.e. $\gamma$ and $\beta$ are dynamically controlled by learnable parameters, $\rho$'s whose sum will be 1 constrained by softmax function. The idea behind using multi-modal adaptive normalization is that various features in the multi-modal domain are correlated. In video synthesis using speech and static image as an input, the video and audio features are related to each other such as optical flow, keypoint heatmap, pitch, energy of speech signals that help in guiding the model to generate expressive realistic videos.

\item Multi-Modal Adaptive normalization opens the non-trivial path to capture the mutual dependence among various domains. Generally, various encoder architectures\cite{Alpher05} are used to convert the various features of multiple domains into latent vectors, and then the concatenated vectors are fed to the decoder to model the mutual dependence and generate the required output. The proposed multi-modal adaptive normalization helps in reducing the number of model parameters required to incorporate the multi-modal mutual dependence into the architecture.

\item Various Experimental and Ablation Study(refer: table~\ref{tab:table55} and Figure~\ref{fig:manres111}) have shown that the proposed normalization is flexible in building the architecture with various networks such as 2DConvolution, partial2D convolution, attention, LSTM, Conv1D for extracting and modeling the mutual information. 

\item Incorporation of multi-modal adaptive normalization for video synthesis using audio and a single image as an input has have shown superior performance on multiple qualitative and quantitative metrics such as SSIM(structural similarity index), PSNR (peak signal to noise ratio), CPBD (image sharpness), WER(word error rate), blinks/sec and LMD(landmark distance).

\end{itemize}

\section{Related Work}

\subsection{Normalization techniques}Normalization is generally applied to improve convergence speed during training. Batch Normalization(BN)\cite{batchnorm} layers were originally designed to accelerate the training of discriminative networks, but now are effective in generative image modeling also. Recently, several alternative normalization schemes have been proposed to extend BN’s effectiveness to recurrent architectures.

Instance Normalization(IN)\cite{isnorm} treats each instance in a mini-batch independently and computes the statistics across only spatial dimensions(H and W). IN aims to make a small change in the stylizing architecture which results in a significant qualitative improvement in the generated images. Adaptive instance normalization(AdaIN)\cite{ADAIN} is a simple extension to IN which adaptively computes the affine parameters from the style input.

Style GAN\cite{stylegan} uses the mapping network to map the input into latent space which then controls the generator using AdaIN\cite{ADAIN}. Each feature map of AdaIN is normalized separately and then scaled and biased using the corresponding scalar components from style features coming from the latent space. Spatially Adaptive Normalization proposed in SPADE\cite{park2019SPADE} takes semantic maps as an input to make learnable parameters i.e.$\gamma$(y) and $\beta$(y) spatially adaptive to it. In U-GAT-IT~\cite{ugatit}, the architecture proposes Adaptive and Layer Instance Instance Normalization which used both normalizations to learn the parameters of the standardization process. 

\cite{WCW} uses the optical flow and structure for motion images as an input to generate the affine parameters, $\gamma$'s and $\beta$'s are multiplied with normalized feature map and fed to the generator to create world consistent videos. 

We have used the features of speech and video domains to generate the affine parameter $\gamma$'s and $\beta$'s in our multi-modal adaptive normalization. The proportion of different parameters are controlled by learnable parameters,$\rho$'s which is fed into softmax function to make the sum equal to 1.We have applied this multi-modal adaptive normalization in the generation of realistic videos given and audio and a static image.

\subsection{Audio to Realistic Video Generation}

The earliest methods for generating videos relied on Hidden Markov Models which captured the dynamics of audio and video sequences. Simons and Cox~\cite{SimonCox} used the Viterbi algorithm to calculate the most likely sequence of mouth shape given the particular utterances. Such methods are not capable of generating quality videos and lack emotions.

\subsubsection{Phoneme and Visemes generation of Videos}
 Phoneme and Visemes based approaches have been used to generate the videos. Real-Time Lip Sync for Live 2D Animation~\cite{Alpher04} has used an LSTM based approach to generate live lip synchronization on 2D character animation.\par

Some of these methods target rigged 3D characters or meshes with predefined mouth blend shapes that correspond to speech sounds ~\cite{inproceedings1,article09,inproceedings3,article10,article11,article12} which primarily focused on mouth motions only and show a finite number of emotions, blinks, facial action units movements.

\subsubsection{Deep learning techniques for Video Generation}

\paragraph{CNN based architectures for audio to video generation}
A lot of work has been done on CNN to generate realistic videos given an audio and static image as input. ~\cite{Chung17b}(Speech2Vid) has used encoder-decoder architecture to generate realistic videos. They have used L1 loss between the synthesized image and the target image. Our approach has used multi-modal adaptive normalization in GAN based architecture to generate realistic videos. 

Synthesizing Obama: Learning Lip Sync from Audio~\cite{article12} is able to generate quality videos of Obama speaking with accurate lip-sync using RNN based architecture. They can generate only a single person video whereas the proposed model can generate videos on multiple images in GAN based approach.

\paragraph{GAN based architectures for audio to video generation}
LumièreNet: Lecture Video Synthesis from Audio~\cite{temporalLoss} generates high-quality, full-pose headshot lecture videos from the instructor’s new audio narration of any length. They have used dense pose~\cite{densepose}, LSTM, variational auto-encoder~\cite{vae} and GANs based approach to synthesize the videos. They are not able to generate the lip-synced videos and the quality of output is low. The proposed method is able to generate lip-synced, expressive videos using audio and a single image as an input. They have used Pix2Pix~\cite{Pix2Pix} for the frame synthesis and we are using a multi-modal adaptive normalization in the generator for frame generation which is able to generate highly expressive and realistic videos.

Temporal Gan~\cite{Temporalgan} and Generating Videos with Scene Dynamics~\cite{SceneDynamics} have done the straight forward adaptation of GANs for generating videos by replacing 2D convolution layers with 3D convolution layers. Such methods are able to capture temporal dependencies but require constant length videos. The proposed model is able to generate videos of variable length with a low word error rate.

Realistic Speech-Driven Facial Animation with GANs(RSDGAN)~\cite{Alpher05} used GAN based approach to produce quality videos. They used identity encoder, context encoder and frame decoder to generate images and used various discriminators to take care of different aspects of video generation. The proposed method has used multi-modal adaptive normalization along with class activation layers and optical flow predictor and keypoint heatmap predictor in the GAN based setting to generate expressive videos.

X2face~\cite{Wiles18} model uses GANs based approach to generate videos given a driving audio or driving video and a source image as an input. The model learns the face embeddings of source frame and driving vectors of driving frames or audio bases which generates the videos. In X2face, the video is processed at 1fps whereas the model generates the video at 25fps. The quality of output video is not good as compared to our proposed method with audio as an input.

The MoCoGAN~\cite{Tulyakov:2018:MoCoGAN} uses RNN based generators with separate latent spaces for motion and content. A sliding window approach is used so that the discriminator can handle variable-length sequences. This model is trained to generate disentangled content and motion vectors such that they can generate audios with different emotions and contents. Our approach uses multi-modal adaptive normalization to generate expressive videos.

Animating Face using Disentangled Audio Representations~\cite{disentangle} has generated the disentangled representation of content and emotion features to generate realistic videos. They have used variational autoencoders~\cite{vae} to learn representation and feed them into GANs based model to generate videos. Audio-driven Facial Reenactment~\cite{neuralhead} used AudioexpressionNet to generate 3D face model. The estimated 3D face model is rendered using the rigid pose observed from the original target image. The proposed method generates 2D videos using multi-modal adaptive normalization in GAN based approach.

\cite{Yi2020AudiodrivenTF} extracts the expression and pose from an audio signal and a 3D face is reconstructed on the target image. The model renders the 3D facial animation into video frames using the texture and lighting information obtained from the input video. Then they fine-tune these synthesized frames into realistic frames using a novel memory-augmented GAN module. The proposed approach uses multi-modal adaptive normalization with predicted optical flow/keypoint heatmap as an input to learn the movements and facial expressions on the target image with audio as an input. \cite{ECCVspeech} have used the L-GAN and T-GAN for motion(landmark) and texture generation. They have used a noise vector for blink generation. MAML is used to generate the videos on an unseen person image. The proposed method has used multi-modal adaptive normalization to generate realistic videos. 

\cite{stochastic} uses variational auto-encoders to generate the video from audio. The generated faces have limited blinks and facial action unit movements. The proposed method generates expressive videos with blinks, cheeks and head movements aspects in it.
\cite{ATVG} uses an audio transformation network (AT-net) for audio to landmark generation and a visual generation network for facial generation. \cite{lipglance} uses audio, identity encoder and a three-stream GAN discriminator for audio, visual and optical flow to generate the lip movement based on input speech. \cite{Zhou2019TalkingFG}  enables arbitrary-subject talking face generation by learning disentangled audio-visual representation through an associative-and-adversarial training process. \cite{Lip2Wav} uses a generator that contains of three blocks: (i) Identity Encoder, (ii) Speech Encoder, and (iii) Face Decoder. It is trained adversarially with Visual quality discriminator and pretrained architecture for lip audio synchronization. \cite{lipglance, Zhou2019TalkingFG, Lip2Wav} are limited to lip movements whereas the proposed method uses multi-modal adaptive normalization to generate different facial action units of an expressive video. \cite{Zhu2020ArbitraryTF} uses Asymmetric Mutual Information Estimator (AMIE) to better express the audio information into generated video in talking face generation. They have AIME to capture the mutual information to learn the cross-modal coherence whereas we have used the multi-modal adaptive normalization to incorporate the multi-modal features into our architecture to generate the expressive videos. \cite{Kumar_2020_CVPR_Workshops} have used deep speech features into the generator architecture with spatially adaptive normalization layers in it along with lip frame discriminator, temporal discriminator and synchronization discriminator to generate realistic videos. They have limited blinks and lip synchronization whereas the proposed method used multi-modal adaptive normalization to capture the mutual relation between audio and video to generate expressive video.

\section{Architectural Design}

Given an arbitrary image and an audio sample, the proposed method is able to generate speech synchronized realistic video on the target face. The proposed method uses multi-modal  adaptive normalization technique to generate the realistic expressive videos. The proposed architecture is GAN-based which consists of a generator and a discriminator.

\begin{figure}[h!]
  \includegraphics[width=\linewidth]{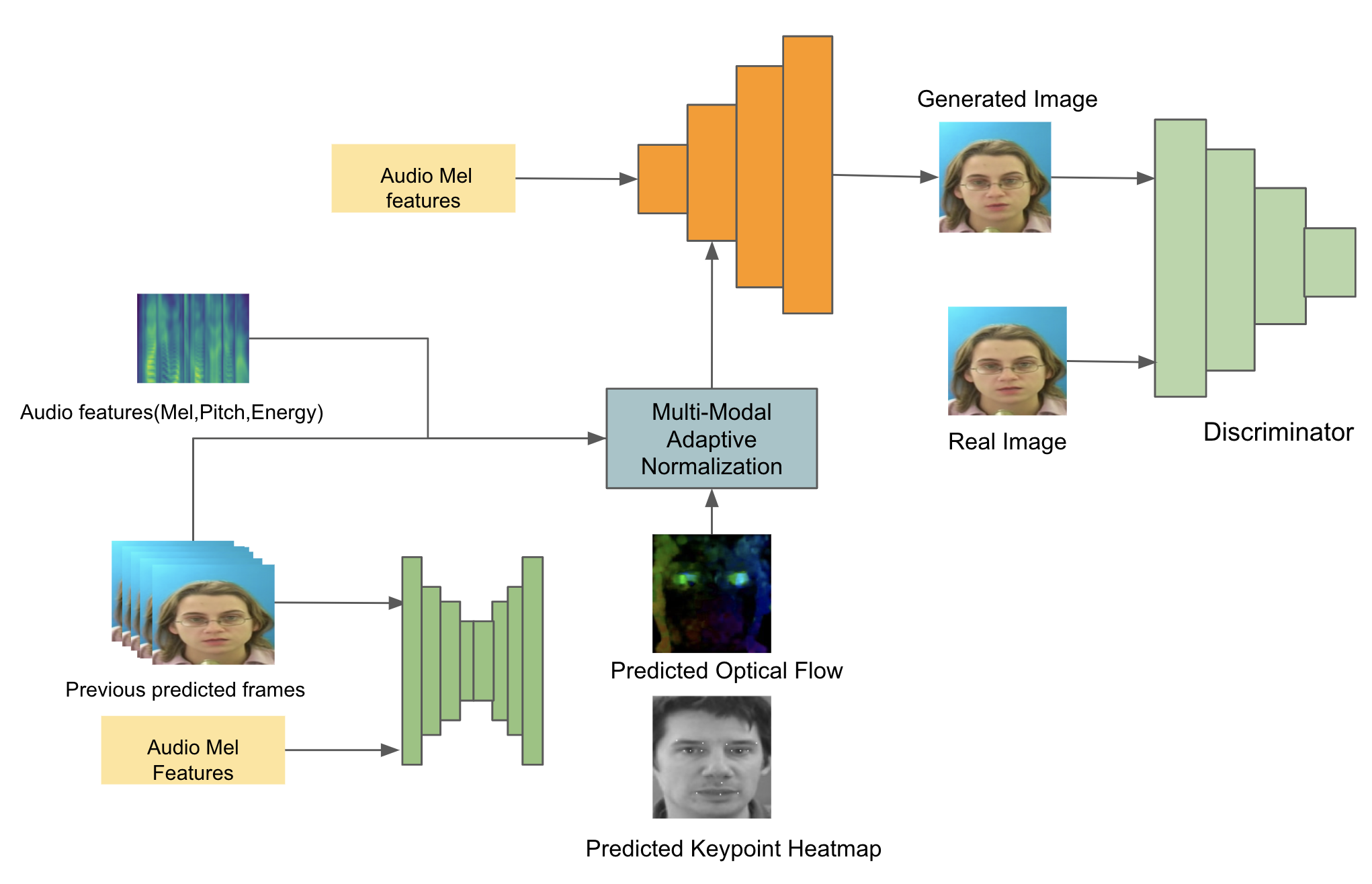}
  \caption{Proposed architecture for Audio to Video synthesis}
  \label{fig:stann}
\end{figure}

\subsubsection{Generator} 
\begin{figure}
  \includegraphics[width=0.8\linewidth]{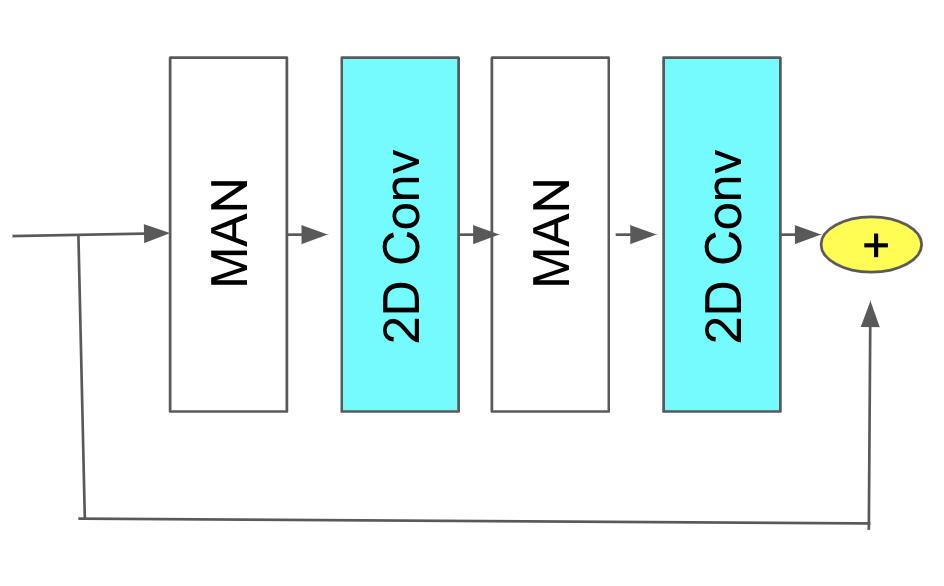}
  \caption{Multi-Modal  Adaptive Normalization based Resnet Block(MAN Resnet Block) . This consists of two MAN block(white block) and two 2D convolution layers(blue block) }
  \label{fig:manres}
\end{figure}

The generator consists of convolution layers with several layers having multi-modal adaptive normalization based resnet block refer Figure ~\ref{fig:manres}. 13 dimensional mel spectrogram features goes to the initial layers of generator and in the subsequent layers, multi-modal adaptive normalization is used to generate expressive videos. The third last layers incorporates the class activation map(CAM)\cite{cam} to focus the generator to learn distinctive features to create expressive videos. The generator uses various architectures namely multi-modal adaptive normalization, CAM and optical flow predictor/keypoint predictor.

\paragraph{KeyPoint Heatmap Predictor} The predictor model is based on Hourglass architecture\cite{hourglass} that, from the input image, estimates K heatmaps $H\textsubscript{K}$ $\epsilon$ [0, 1]H×W, one for each keypoint, each of which represents the contour of a specific facial part, e.g., upper left eyelid and nose bridge. It captures the spatial configuration of all landmarks, and hence it captures pose, expression and shape information. We have used pretrained model\cite{facek} to calculate the ground truth of heatmap and have applied mean square error loss between predicted heatmaps and ground truth. In the experiments, we have used the 15 channel heatmaps and input and output sizes are (15,96,96). We have done the joint training of keypoint predictor architecture along with the generator architecture and fed the output of keypoint predictor architecture in the multi-modal adaptive normalization network to learn the affine parameters and have optimized it with mean square error loss with the output of pretrained model\cite{facek}. The input of the keypoint predictor model is the previous 5 frames along with 256 audio mel spectrogram-features which are concatenated along the channel axis.

\begin{figure}
  \includegraphics[width=\linewidth]{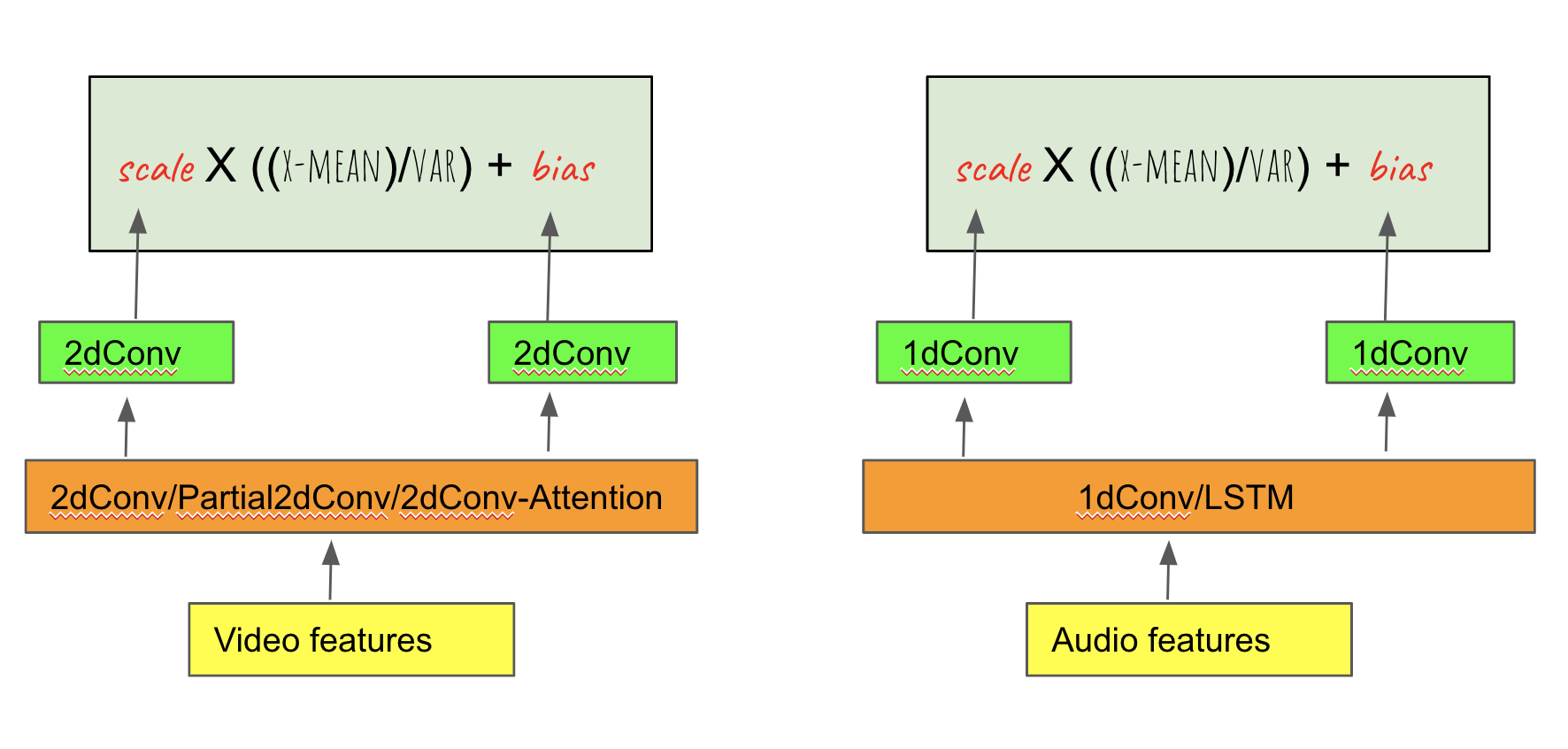}
  \caption{Single block of multi modal adaptive normalization(MAN) showing the architectural design while incorporating video and audio features into it.}
  \label{fig:manres111}
\end{figure}

\paragraph{Optical Flow Predictor} The architecture is based on UNET \cite{Unet} to predict the optical flow of the next frame. We are giving the previous frames and current audio mel-spectrogram as an input to the model with KL loss and reconstruction loss. The pretrained model is then used in the generator to calculate the affine parameters. The input of the optical flow is previous 5 frames along with 256 audio mel spectrogram-features and is jointly trained along with generator architecture and is optimized with mean square loss with the actual optical loss.

\paragraph{Multi-Modal Adaptive Normalization} - We have used the pitch, energy and audio mel-spectrogram features(AMF) from audio domain \& static image and optical flow(OF)/facial keypoints heatmap(KH) features from video domain in the normalization to compute the different affine parameters in multi-modal adaptive normalization setup.Multi-modal adaptive normalization gives the flexibility of using various architectures namely 2D convolution, Partial Convolution and attention model for video related features and 1D convolution and LSTM layer for audio features as shown in  Figure ~\ref{fig:manres111}. The parameter $\rho$'s is used to combine these parameters (Equation~\eqref{conv1}). The value of $\rho$'s is constrained to the range of [0, 1] by using softmax function (Equation~\eqref{rhosum}). Figure ~\ref{fig:manres} shows MAN resnet block to create the framework which consists of 2 MAN block and 2 layers of 2D convolution with the kernel size of 3*3. We have used 8 MAN resnet block in the generator architecture with varying channel size.

\begin{dmath}
\label{conv1}
    y = \rho\textsubscript{1}(\gamma\textsubscript{Image} x\textsubscript{IN} + \beta\textsubscript{Image}) + \rho\textsubscript{2}(\gamma\textsubscript{OF/KH} x\textsubscript{IN} + \beta\textsubscript{OF/KH})+ \rho\textsubscript{3}(\gamma\textsubscript{AMF} x\textsubscript{IN} + \beta\textsubscript{AMF}) +\rho\textsubscript{4}(\gamma\textsubscript{pitch} x\textsubscript{IN} + \beta\textsubscript{pitch}) + \rho\textsubscript{5}(\gamma\textsubscript{energy} x\textsubscript{IN} + \beta\textsubscript{energy})
\end{dmath}
 \begin{equation}
 \label{rhosum}
   \rho\textsubscript{1}+\rho\textsubscript{2}+\rho\textsubscript{3}+\rho\textsubscript{4}+\rho\textsubscript{5} = 1
\end{equation}

\paragraph{Class Activation Map(CAM)\cite{cam} based layer} This layer is employed on third last layer of generator to capture the global and local features of face. In Class Activation Map, we have done the concatenation of adaptive average pooling and adaptive max pooling of feature map to create the CAM features which captures global and local attention map, that helps the generator to focus on the image regions that are more discriminative such as eyes, mouth and cheeks.

\subsubsection{Multi scale Frame discriminator} We have used multi-scale frame discriminator \cite{wang2018pix2pixHD} to distinguish the fake and real image at finer and coarser level. The class activation map based layer is also used to distinguish the real or fake image by visualizing local and global attention maps.We have applied the adversarial loss(Equation~\eqref{camloss})on the information from the CAM output, n\textsubscript{D\textsubscript{t}} at different scale of the discriminator so that it will help the generator and discriminator to focus on local and global features and helps in generating more realistic image.

 \begin{equation}
 \label{camloss}
    L\textsubscript{cam} = E\textsubscript{y$\sim$P\textsubscript{t}}[\log(n\textsubscript{D\textsubscript{t}}(y))] + E\textsubscript{x$\sim$P\textsubscript{s}}[\log(D(1-n\textsubscript{D\textsubscript{t}}(G(x))))] 
\end{equation}

 \begin{table*}[h!]
  \begin{center}

    \begin{tabular}{c|c|c|c|c|c|c|c|c} 
      \textbf{Method} & \textbf{SSIM$\uparrow$} & \textbf{PSNR$\uparrow$} & \textbf{CPBD$\uparrow$} & \textbf{WER$\downarrow$} & \textbf{ACD-C$\downarrow$} & \textbf{ACD-E$\downarrow$} & \textbf{blinks/sec} &\textbf{LMD$\downarrow$}\\
      \hline
      FOMM(GRID)\cite{FOMM}& 0.833 & 26.72 & 0.214 & 38.21  & 0.004 &0.088& 0.56 & \textbf{0.718} \\
       OneShotA2V(GRID)\cite{Kumar_2020_CVPR_Workshops}& 0.881 & 28.571 & 0.262 &27.5 & 0.005 & 0.09 & - & -  \\
       RSDGAN(GRID)\cite{Alpher05}  & 0.818 & 27.100 & 0.268 & \textbf{23.1} & - & 1.47x10\textsuperscript{-4}&0.39 & - \\
       Speech2Vid(GRID)\cite{Chung17b}   & 0.720 & 22.662 & 0.255 & 58.2 & 0.007 & 1.48x10\textsuperscript{-4}&- & - \\
       ATVGnet(GRID)\cite{ATVG}  & 0.83 & \textbf{32.15} & - & - & - & - & - & 1.29\\
       X2face(GRID)\cite{Wiles18} & 0.80 & 29.39 & - & - & - & - & -& 1.48 \\
       LipSyncglance(GRID)\cite{lipglance} & 0.76 & 29.33 & - & - & - & - & - & 1.59\\
       AMIE(GRID)\cite{Zhu2020ArbitraryTF} & \textbf{0.97} & 31.01 & - & - & - & - & - & 0.78 \\
       CascadedGAN(GRID)\cite{ECCVspeech} & 0.81& 27.1 & 0.26 & 23.1 & - & 1.47x10\textsuperscript{-4} & 0.45 & -\\
       MAN-optical(GRID) & 0.908& 29.78 & \textbf{0.272} & 23.7 & 0.005 & 0.008 & 0.45 & 0.77 \\
       MAN-keypoint(GRID) & 0.887 & 29.01 & 0.269& 25.2 & 0.006& 0.010 & 0.48 & 0.80 \\
       \hline
       FOMM(CREMA-D)\cite{FOMM}& 0.654 & 20.74 & 0.186 &  NA & 0.007 &0.12&- &1.041 \\
       OneShotA2V(CREMA-D)\cite{Kumar_2020_CVPR_Workshops}& 0.773 & 24.057 & 0.184 &NA& 0.006 & 0.96 &- &- \\
       RSDGAN(CREMA-D)\cite{Alpher05}  & 0.700 & 23.565 & 0.216 & NA & - & 1.40x10\textsuperscript{-4}&- & -\\
       Speech2Vid(CREMA-D)\cite{Chung17b}    & 0.700 & 22.190 & 0.217 & NA & 0.008 & 1.73x10\textsuperscript{-4} & - &-\\
       MAN-optical(CREMA-D) & 0.826& 27.723 & 0.224 & NA & 0.004 & 0.082 & - & 0.592\\
       MAN-keypoint(CREMA-D) & \textbf{0.84}1 & \textbf{28.01} & \textbf{0.228} & NA & \textbf{0.003} & \textbf{0.079} & - & \textbf{0.51} \\
       \hline
       FOMM(lombard)\cite{FOMM}& 0.804 & 22.97 & 0.381 &  NA  & 0.003 &0.078& 0.37 &1.09  \\
      OneShotA2V(lombard)\cite{Kumar_2020_CVPR_Workshops} & \textbf{0.922} & \textbf{28.978} & \textbf{0.453} & NA &0.002 &0.064 & - &-\\
    Speech2Vid(lombard)\cite{Chung17b} & 0.782 & 26.784 & 0.406 & NA &0.004 &0.069 & - & -\\
       MAN-optical(lombard) & 0.895& 26.94 & 0.43 & NA & 0.001 & 0.048 & 0.21 & 0.588 \\
       MAN-keypoint(lombard) & 0.911 & 27.45 & 0.40 & NA & 0.001 & 0.046 & 0.31 & 0.563 \\
       \hline
       MAN-keypoint(voxceleb2) & 0.732 & 22.41 & 0.126 & NA & 0.004 & 0.088 & - & 0.47\\
    \end{tabular}
    \vspace {0.25\baselineskip}
    \caption{Comparision of the proposed method(MAN-keypoint and MAN-optica) with other previous works for GRID, GRID lombard, CREMA-D and voxceleb datasets for SSIM, PSNR, CPBD, WER, ACD , blinks/sec and LMD by calculating cosine distance(ACD-C)(should be 0.02 and below) and euclidean distance(ACD-E)(should be 0.2 and below).}
    \label{tab:table29}
  \end{center}
\end{table*}

\section{Experiments}
\subsection{Implementation Details}

\paragraph{Datasets and PreProcessing Steps} 
We have used the GRID ~\cite{Alpher03}, LOMBARD GRID~\cite{gridlombard}, Crema-D\cite{crema} and VoxCeleb2\cite{voxceleb} datasets for the experiments and evaluation of different metrics. Videos are processed at 25fps and frames are resized into 256X256 size and audio features are processed at 16khz. The ground truth of optical flow is calculated used farneback optical flow algorithm\cite{farneback}. To extract the keypoint heatmaps, we have used the pretrained hourglass face keypoint detection \cite{facek}. Every audio frame is centered around a single video frame. To do that, zero padding is done before and after the audio signal and use the following formula for the stride.
 \begin{align*}
      stride = \frac{\text{audio sampling rate}}{\text{video frames per sec}}
\end{align*}
We extract the pitch, F0 using using PyWorldVocoder\cite{pyvoc} from the raw waveform with the frame size of 1024 and hop size of 256 sampled at 16khz to obtain the pitch of each frame and compute the L2-norm of the amplitude of each STFT frame as the energy. We quantize the F0 and energy of each frame to 256 possible values and encode them into a sequence of one-hot vectors as p and e respectively and then feed the value of p, e and 256 dimensional mel spectrogram features in the proposed normalization method.

 \subsection{Metrics}
To quantify the quality of the final generated video, we use the following metrics. PSNR(Peak Signal to Noise Ratio), SSIM(Structural Similarity Index), CPBD(Cumulative Probability Blur Detection), ACD(Average Content Distance) and KID(Kernel Inception Distance). PSNR, SSIM, and CPBD measure the quality of the generated image in terms of the presence of noise, perceptual degradation, and blurriness respectively. ACD~\cite{Tulyakov:2018:MoCoGAN} is used for the identification of the speaker from the generated frames by using OpenPose~\cite{cao2018openpose}. Along with image quality metrics, we also calculate WER(Word Error Rate) using pretrained LipNet architecture~\cite{assael2016lipnet}, Blinks/sec using ~\cite{Authors14} and LMD(Landmark Distance)~\cite{lmd} to evaluate our performance of speech recognition, eye-blink reconstruction and lip reconstruction respectively. 

\paragraph{Training and Inference} Our model is implemented in pytorch and takes approximately 7 days to run on 4 Nvidia V100 GPUs for training. In the training stage, the model is trained with multi-scale frame discriminator with loss :
\begin{equation}
    L\textsubscript{GAN}(G,D) = E\textsubscript{x\textsubscript{d}}[\log(D(x))] + E\textsubscript{x\textsubscript{d}}[\log(D(1-G(z)))] 
\end{equation}
Reconstruction loss~\cite{RLoss}, RL is used on the lower half of the image to improve the reconstruction in the mouth area between actual $R\textsubscript{n}$ and predicted image $G\textsubscript{n}$.
\begin{equation}
    L\textsubscript{RL} = \sum_{n\epsilon [0,W]*[H/2,H]}^{}(R\textsubscript{n} - G\textsubscript{n})
\end{equation}
Perceptual Loss~\cite{PerceptualLoss}, Feature-matching Loss~\cite{wang2018pix2pixHD} , CAM loss(refer Equation~\eqref{camloss}) and Key-point predictor/optical flow based mean square error loss are also used to ensure generation of natural-looking, high-quality frames. We have taken the Adam optimizer~\cite{Adam} with learning rate = 0.002 and $\beta_1$= 0.0 and $\beta_2$ = 0.90 for the generator and discriminators.

\subsection{Implementation Results}
\subsubsection{Quantitative Results}
Table~\ref{tab:table29} compares the proposed method with its competitors and shows better SSIM, PSNR, CPBD, Word error rate(WER), blinks/sec and LMD on GRID\cite{Alpher03}, Crema-D\cite{crema}, Grid-lombard\cite{gridlombard} and Voxceleb2\cite{voxceleb} datasets, suggesting highly expressive and realistic video synthesis.

\subsubsection{Qualitative Results}
\begin{figure}[h!]
  \includegraphics[width=\linewidth]{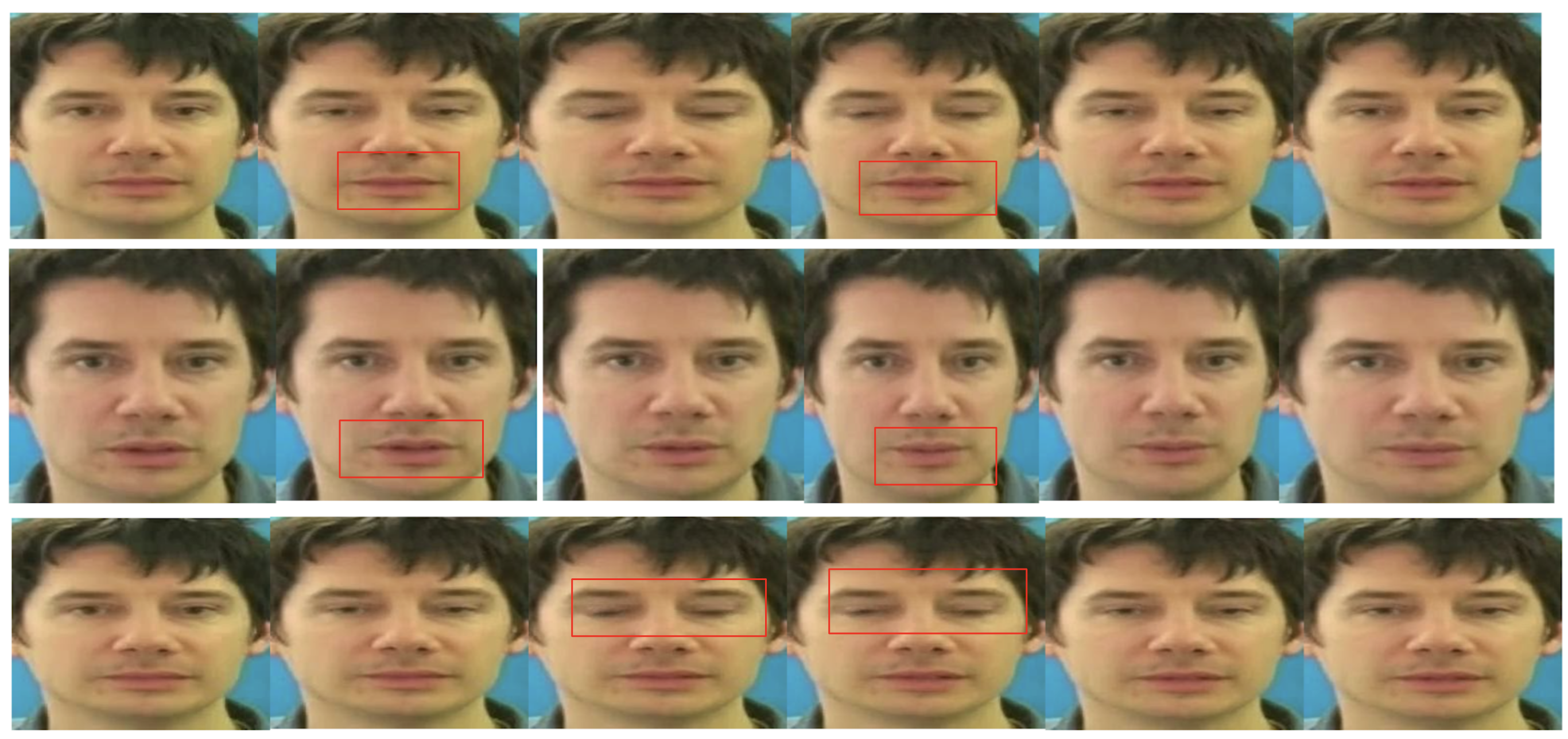}
  \caption{Top: The Speaker speaking the word 'bin' , Middle : The Speaker speaking the word 'please', Bottom: The speaker blinking his eyes}
  \label{fig:eval}
\end{figure}

\begin{figure}[h!]
  \includegraphics[width=\linewidth]{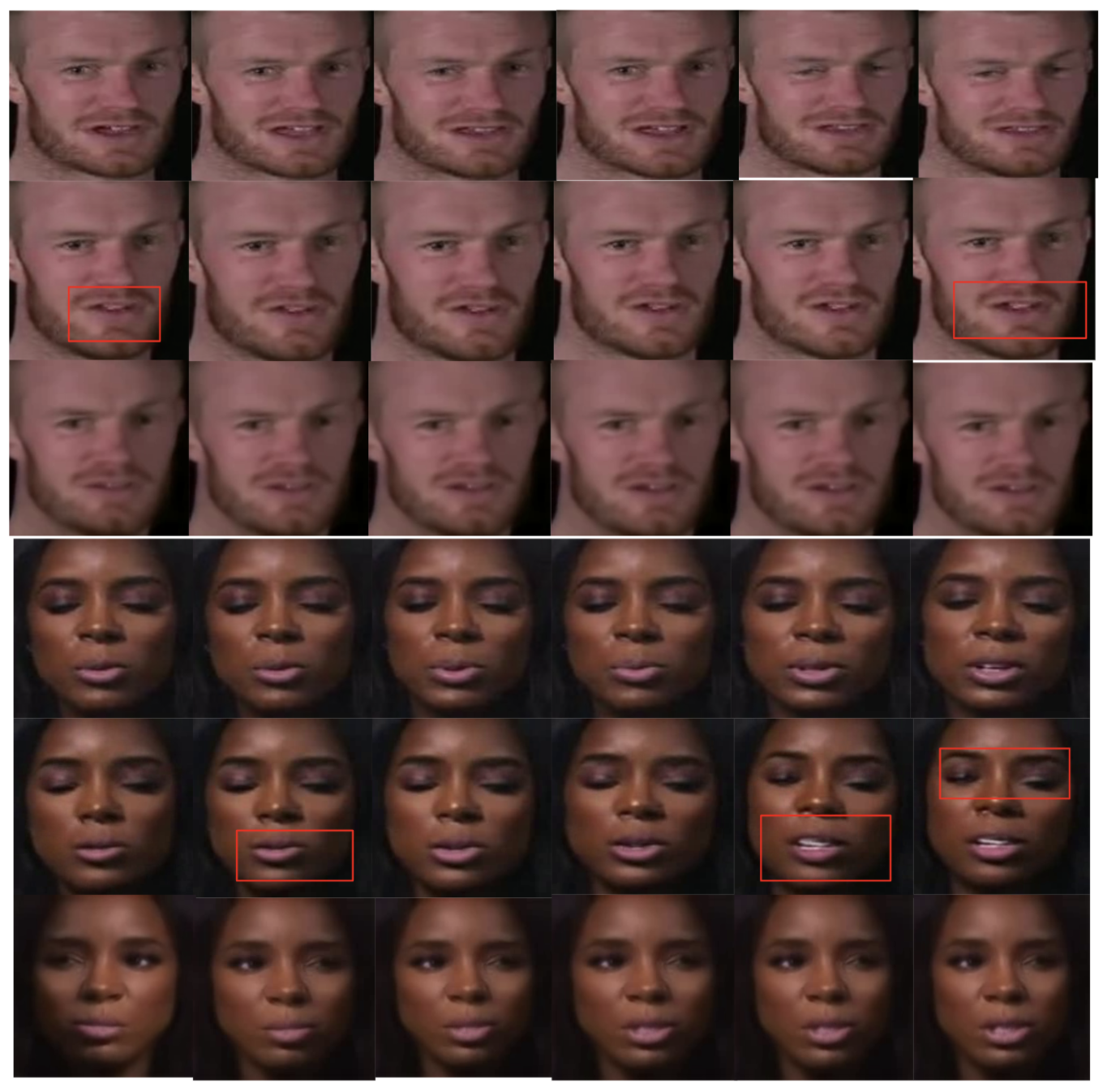}
  \caption{Top: Actual frames of voxceleb2\cite{voxceleb} dataset , Middle : Predicted frames from proposed method, Bottom: Predicted frame from \cite{Zhu2020ArbitraryTF}}
  \label{fig:frames3}
\end{figure}

 \paragraph{Expressive aspect}
 Figure ~\ref{fig:eval} displays the lip synchronized frames of speaker speaking the word 'bin' and 'please' as well as the blinking of the eyes. Figure ~\ref{fig:frames3} shows the comparison of proposed model with previous work \cite{Zhu2020ArbitraryTF} where the proposed model show better image reconstruction and lip synchronization. 

\begin{figure}[h!]
  \includegraphics[width=\linewidth]{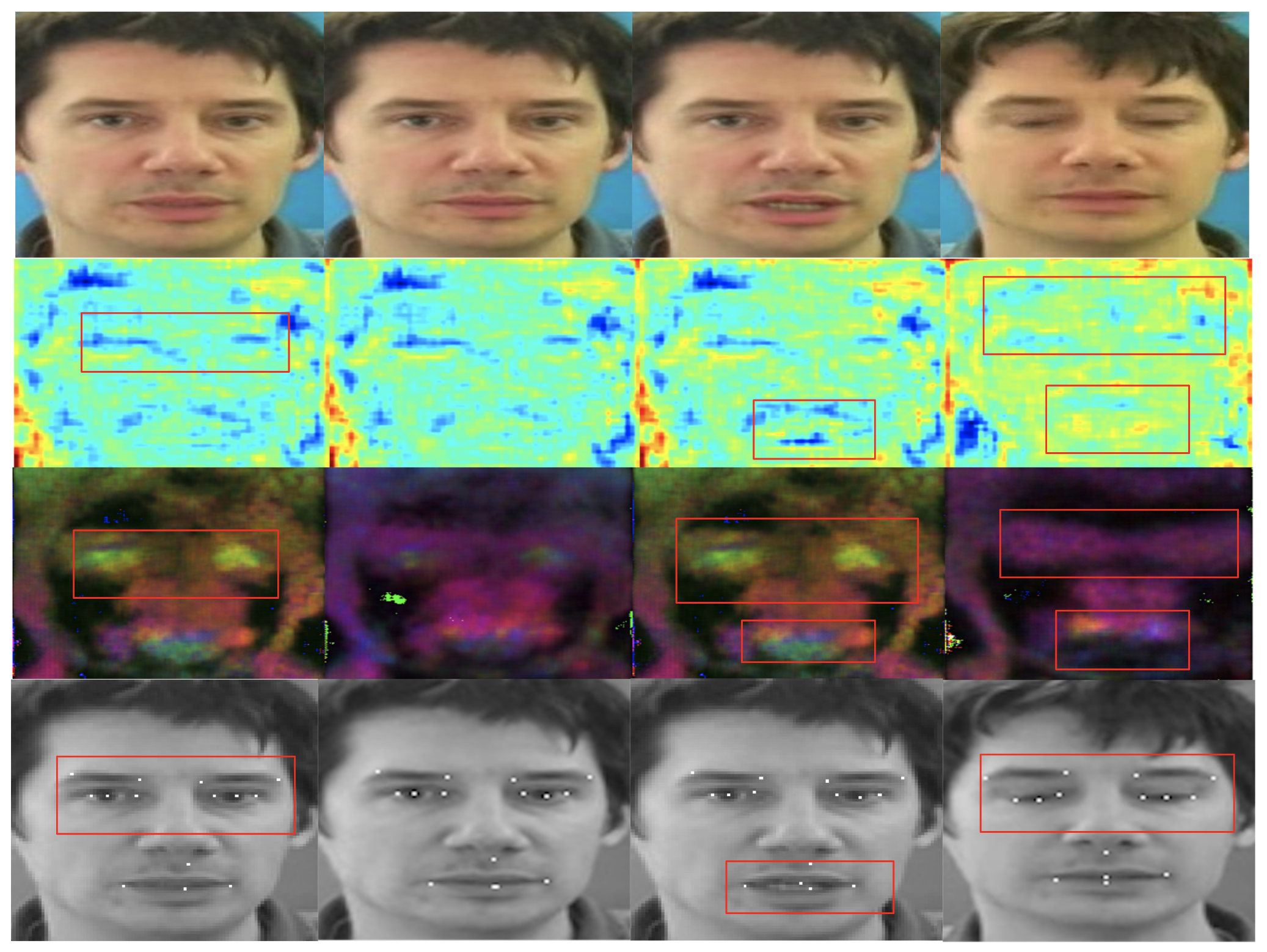}
  \caption{Top: The Speaker with different expressions, Middle1 : CAM based attention map, Middle2: Predicted optical flow from the optical flow generator architecture,Bottom: Predicted Key-points from Key-point predictor architecture}
  \label{fig:arch}
\end{figure}

\paragraph{Architecture Analysis}
  Figure ~\ref{fig:arch} shows the optical flow map and Class activation based heatmap at different expression of the speaker while speaking. The optical flow map has different color while speaking and opening of eyes as compared to closing of mouth and blinking of eyes. The CAM based heatmap shows the attention regions in the heatmap which captures the local as well as global features while video generation. Bottom part of the figure shows the keypoints from predicted heatmap from keypoint predictor calculated using the max operator to find the coordinates of the maximum value in each predicted heatmap(15,96,96).

\begin{figure}[h!]
  \includegraphics[width=0.8\linewidth]{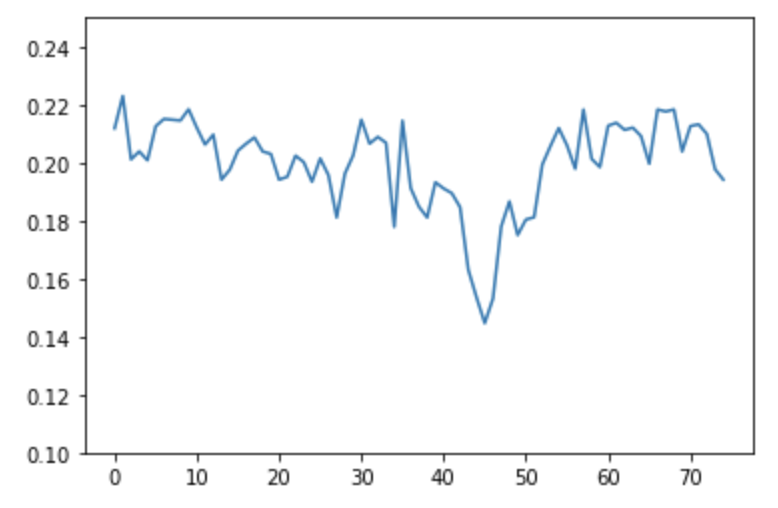}
  \caption{Eye Aspect Ratio of the predicted video with 75 frames on GRID dataset}
  \label{fig:eyeeval}
\end{figure}

\paragraph{Eye Blinks and head movement}
 The average human blink rate of 0.28 blinks/second, especially when considering that the blink rate increases to 0.4 blinks/second during conversation.Figure ~\ref{fig:eyeeval} shows the sharp decline in the eye aspect ratio ~\cite{Authors14} at the centre which justifies the generation of blinks in the predicted videos. Table~\ref{tab:table29} shows the blinks/sec of 0.45 on GRID dataset.

\begin{figure}
  \includegraphics[width=\linewidth]{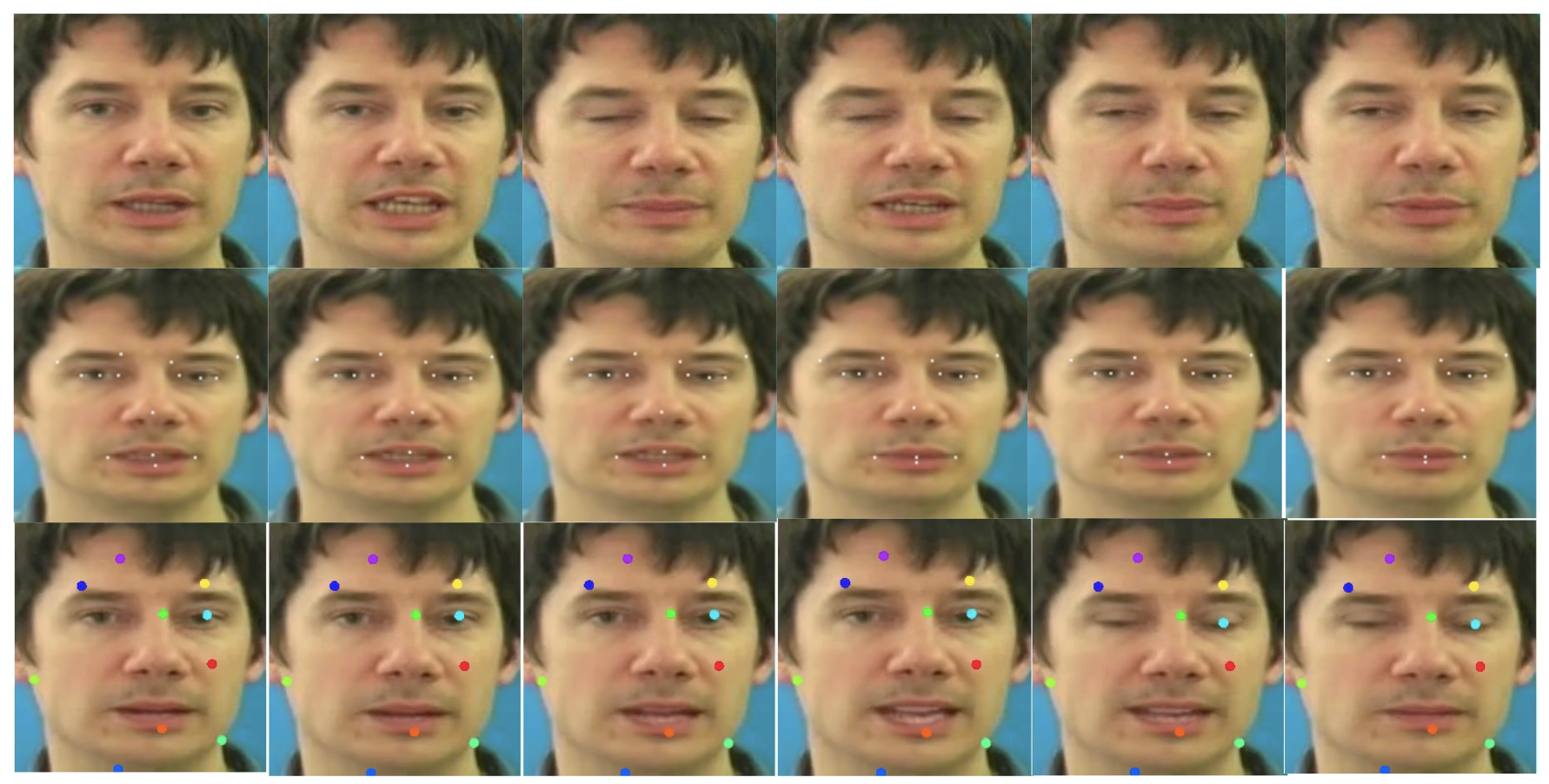}
  \caption{Top: Actual frames of speaker of GRID dataset. Middle: Predicted frames from proposed method with keypoints predicted from keypoint predictor. Bottom: Predicted frames from FOMM,\cite{FOMM} method}
  \label{fig:fomm}
\end{figure}

\paragraph{Comparison with Video to Video Synthesis Architecture}
We have compared the proposed method with first order motion model for image animation, (FOMM)\cite{FOMM} on grid dataset which generates the video sequences so that an object in a source image is animated according to the motion of a driving video. The comparison is done to see how effectively driving audio signals instead of driving video helps in reconstructing the expressive video as shown in Figure ~\ref{fig:fomm}. Table~\ref{tab:table29} compares the various metrics between FOMM and the proposed model and shown better image reconstruction metrics(SSIM, PSNR, CPBD,LMD) and WER but FOMM has more blinks/sec as compared to the proposed method. The reason for  better WER is limited number of utterances in GRID dataset and faster speaking style of the speaker which the proposed method is better able to capture as compared to FOMM.

\subsection{Ablation Study}
\paragraph{Network Analysis in Multi Modal Adaptive Normalization} We have done the ablation study on three architectures namely 2D convolution,Partial 2DConvolution\cite{liu2018partialpadding,liu2018partialinpainting} and 2D convolution+Efficient channel Attention(ECA)\cite{eca} for extracting video features and two architectures namely 1D convolution and LSTM for audio features as shown in Figure ~\ref{fig:manres111} to study its effect on multi-modal adaptive normalization  with optical flow predictor  in the proposed method. Table~\ref{tab:table55} shows that 2DConv+ECA+LSTM  has improved the reconstruction metrics such as SSIM, PSNR and CPBD as well as word error rate and blinks/sec as compared to other networks. The image quality reduced with use of partial2D convolution which demonstrates that since the predicted optical flow is dense, holes in the optical flow has some spatial relation with other regions which is better captured by other networks.

\begin{table}[h!]

 
\scalebox{0.7}{
    \begin{tabular}{c|c|c|c|c|c} 
      \textbf{Method} & \textbf{SSIM$\uparrow$} & \textbf{PSNR$\uparrow$} & \textbf{CPBD$\uparrow$} & \textbf{blinks/sec} & \textbf{WER$\downarrow$}\\
      \hline
       2DConv+1dConv  & 0.875 & 28.65 & 0.261 & 0.35 & 25.6\\
       Partial2DConv+1dConv & 0.803 & 28.12 &  0.256 & 0.15 & 29.4\\
       2DConv+ECA+1dConv   & 0.880 & 29.11 & 0.263 & 0.42 & 23.9\\
       2DConv+LSTM  & 0.896 & 29.25 & 0.086 & 0.260 & 24.1\\
       Partial2DConv+LSTM & 0.823 & 28.12 &  0.258 & 0.12 & 28.3\\
       2DConv+ECA+LSTM   & 0.908 & 29.78 & 0.272 & 0.45 & 23.7\\
       
    \end{tabular}
    }
    \vspace {0.25\baselineskip}
    \caption{Ablation study of different networks of multi-modal adaptive normalization on Grid dataset }
    \label{tab:table55}
\end{table}

\paragraph{Incremental effect of multi modal adaptive normalization}
We studied the incremental effect of multi modal adaptive normalization of the proposed model with optical flow predictor(OFP) and 2DConv+ECA+LSTM combination in multi modal attention normalization on GRID dataset. Table 3 shows the impact of addition of mel-spectrogram features, pitch, predicted optical flow in the multi modal adaptive normalization. Base model consists generator and discriminator architecture with static image in the adaptive normalization.

\begin{table}[h!]
\scalebox{0.7}{
    \begin{tabular}{c|c|c|c|c|c} 
      \textbf{Method} & \textbf{SSIM$\uparrow$} & \textbf{PSNR$\uparrow$} & \textbf{CPBD$\uparrow$} & \textbf{blinks/sec} & \textbf{WER$\downarrow$}\\
      \hline
       Base Model(BM)  & 0.776 & 27.99 &  0.213 & 0.02 & 57.9\\
       BM + OFP+mel    & 0.878 & 28.43 &  0.244 & 0.38 & 27.4\\
       BM + OFP+mel+pitch  & 0.881& 28.57 & 0.264 & 0.41 & 24.1\\
       BM+OFP+mel+pitch+energy & 0.908 & 29.78 & 0.272 & 0.45 & 23.7\\
    \end{tabular}
    }
    \vspace {0.25\baselineskip}
    \caption{Incremental study Multi Modal Adaptive Normalization on Grid dataset  }
    \label{tab:incr}
\end{table}

\section{Conclusions and Future Work}

In this paper, we propose a novel multi-modal adaptive normalization for generating expressive video from audio and a single image as an input. We have seen how multi-modal adaptive normalization captures the mutual relationship across the domains to generate realistic expressive videos. We have also shown how this novel architecture is flexible in using different kinds of networks such as 2D convolution, partial convolution, attention, LSTM, etc. Experimental evaluation of various datasets demonstrates superior performance compared to other related works on multiple quantitative and qualitative metrics.

In the future, we plan to expand this novel architecture on other multi-modal applications such as image captioning, 3D video synthesis, text to speech, etc. This approach of normalization will open the novel path of capturing mutual information in an efficient way in terms of model parameters and training time.


\begin{thebibliography}{10}\itemsep=-1pt

\bibitem{Alpher03}
FirstName Alpher and FirstName Fotheringham-Smythe.
\newblock Frobnication revisited.
\newblock {\em Journal of Foo}, 13(1):234--778, 2003.

\bibitem{Alpher04}
FirstName Alpher, FirstName Fotheringham-Smythe, and FirstName Gamow.
\newblock Can a machine frobnicate?
\newblock {\em Journal of Foo}, 14(1):234--778, 2004.

\bibitem{assael2016lipnet}
Yannis~M Assael, Brendan Shillingford, Shimon Whiteson, and Nando de Freitas.
\newblock Lipnet: End-to-end sentence-level lipreading.
\newblock {\em GPU Technology Conference}, 2017.

\bibitem{LN}
Jimmy Ba, Jamie Kiros, and Geoffrey Hinton.
\newblock Layer normalization.
\newblock 07 2016.

\bibitem{crema}
Houwei Cao, David Cooper, Michael Keutmann, Ruben Gur, Ani Nenkova, and Ragini
  Verma.
\newblock Crema-d: Crowd-sourced emotional multimodal actors dataset.
\newblock {\em IEEE transactions on affective computing}, 5:377--390, 10 2014.

\bibitem{cao2018openpose}
Zhe Cao, Gines Hidalgo, Tomas Simon, Shih-En Wei, and Yaser Sheikh.
\newblock Open{P}ose: realtime multi-person 2{D} pose estimation using {P}art
  {A}ffinity {F}ields.
\newblock In {\em arXiv preprint arXiv:1812.08008}, 2018.

\bibitem{lipglance}
Lele Chen, Zhiheng Li, Ross Maddox, Zhiyao Duan, and Chenliang Xu.
\newblock Lip movements generation at a glance.
\newblock 07 2018.

\bibitem{lmd}
Lele Chen, Zhiheng Li, Ross~K. Maddox, Zhiyao Duan, and Chenliang Xu.
\newblock Lip movements generation at a glance.
\newblock 2018.

\bibitem{ATVG}
Lele Chen, Ross Maddox, Zhiyao Duan, and Chenliang Xu.
\newblock Hierarchical cross-modal talking face generationwith dynamic
  pixel-wise loss, 05 2019.

\bibitem{Chung17b}
Joon~Son Chung, Amir Jamaludin, and Andrew Zisserman.
\newblock You said that?
\newblock In {\em British Machine Vision Conference}, 2017.

\bibitem{voxceleb}
J.~S. Chung, A. Nagrani, and A. Zisserman.
\newblock Voxceleb2: Deep speaker recognition.
\newblock In {\em INTERSPEECH}, 2018.

\bibitem{ECCVspeech}
Sanjana~Sinha Dipanjan~Das, Sandika~Biswas and Brojeshwar Bhowmick.
\newblock Speech-driven facial animation using cascaded gans for learning of
  motion and texture.
\newblock 10 2019.

\bibitem{article10}
Pif Edwards, Chris Landreth, Eugene Fiume, and Karan Singh.
\newblock Jali: an animator-centric viseme model for expressive lip
  synchronization.
\newblock {\em ACM Transactions on Graphics}, 35:1--11, 07 2016.

\bibitem{facek}
https://github.com/raymon-tian/hourglass-facekeypoints-detection
  facial~keypoint detection.

\bibitem{farneback}
Gunnar Farnebäck.
\newblock Two-frame motion estimation based on polynomial expansion.
\newblock volume 2749, pages 363--370, 06 2003.

\bibitem{densepose}
Riza Guler, Natalia Neverova, and Iasonas Kokkinos.
\newblock Densepose: Dense human pose estimation in the wild.
\newblock pages 7297--7306, 06 2018.

\bibitem{ADAIN}
Xun Huang and Serge Belongie.
\newblock Arbitrary style transfer in real-time with adaptive instance
  normalization.
\newblock pages 1510--1519, 10 2017.

\bibitem{batchnorm}
Sergey Ioffe and Christian Szegedy.
\newblock Batch normalization: Accelerating deep network training by reducing
  internal covariate shift.
\newblock 02 2015.

\bibitem{Pix2Pix}
Phillip Isola, Jun-Yan Zhu, Tinghui Zhou, and Alexei Efros.
\newblock Image-to-image translation with conditional adversarial networks.
\newblock pages 5967--5976, 07 2017.

\bibitem{PerceptualLoss}
Alexandre~Alahi Justin~Johnson and Li Fei-Fei.
\newblock Perceptual losses for real-time style transfer and super-resolution.
\newblock 2016.

\bibitem{article09}
Tero Karras, Timo Aila, Samuli Laine, Antti Herva, and Jaakko Lehtinen.
\newblock Audio-driven facial animation by joint end-to-end learning of pose
  and emotion.
\newblock {\em ACM Transactions on Graphics}, 36:1--12, 07 2017.

\bibitem{temporalLoss}
Byung-Hak Kim and Varun Ganapath.
\newblock Lumièrenet: Lecture video synthesis from audio.
\newblock 2016.

\bibitem{ugatit}
Junho Kim, Minjae Kim, Hyeon-Woo Kang, and Kwanghee Lee.
\newblock U-gat-it: Unsupervised generative attentional networks with adaptive
  layer-instance normalization for image-to-image translation, 07 2019.

\bibitem{Adam}
Diederik Kingma and Jimmy Ba.
\newblock Adam: A method for stochastic optimization.
\newblock {\em International Conference on Learning Representations}, 12 2014.

\bibitem{Kumar_2020_CVPR_Workshops}
Neeraj Kumar, Srishti Goel, Ankur Narang, and Mujtaba Hasan.
\newblock Robust one shot audio to video generation.
\newblock In {\em Proceedings of the IEEE/CVF Conference on Computer Vision and
  Pattern Recognition (CVPR) Workshops}, June 2020.

\bibitem{Authors14}
FirstName LastName.
\newblock The frobnicatable foo filter, 2014.
\newblock Face and Gesture submission ID 324. Supplied as additional material
  {\tt fg324.pdf}.

\bibitem{RLoss}
Yanchun Li, Nanfeng Xiao, and Wanli Ouyang.
\newblock Improved generative adversarial networks with reconstruction loss.
\newblock {\em Neurocomputing}, 323, 10 2018.

\bibitem{liu2018partialinpainting}
Guilin Liu, Fitsum~A. Reda, Kevin~J. Shih, Ting-Chun Wang, Andrew Tao, and
  Bryan Catanzaro.
\newblock Image inpainting for irregular holes using partial convolutions.
\newblock In {\em The European Conference on Computer Vision (ECCV)}, 2018.

\bibitem{liu2018partialpadding}
Guilin Liu, Kevin~J. Shih, Ting-Chun Wang, Fitsum~A. Reda, Karan Sapra, Zhiding
  Yu, Andrew Tao, and Bryan Catanzaro.
\newblock Partial convolution based padding.
\newblock In {\em arXiv preprint arXiv:1811.11718}, 2018.

\bibitem{WCW}
Arun Mallya, Ting-Chun Wang, Karan Sapra, and Ming-Yu Liu.
\newblock World-consistent video-to-video synthesis, 07 2020.

\bibitem{article11}
Wesley Mattheyses and Werner Verhelst.
\newblock Audiovisual speech synthesis: An overview of the state-of-the-art.
\newblock {\em Speech Communication}, 66, 11 2014.

\bibitem{disentangle}
Gaurav Mittal and Baoyuan Wang.
\newblock Animating face using disentangled audio representations, 2019.

\bibitem{gridlombard}
Ricard Marxer-Jon~Barker Najwa~Alghamdi, Steve~Maddock and Guy~J. Brown.
\newblock A corpus of audio-visual lombard speech with frontal and profile
  view, the journal of the acoustical society of america 143, el523 (2018);
  https://doi.org/10.1121/1.5042758, 2018.

\bibitem{batchins}
Hyeonseob Nam and Hyo-Eun Kim.
\newblock Batch-instance normalization for adaptively style-invariant neural
  networks, 05 2018.

\bibitem{stylegan}
Dmitry Nikitko.
\newblock stylegan-encoder. https://github. com/puzer/stylegan-encoder, 2019.

\bibitem{park2019SPADE}
Taesung Park, Ming-Yu Liu, Ting-Chun Wang, and Jun-Yan Zhu.
\newblock Semantic image synthesis with spatially-adaptive normalization.
\newblock In {\em Proceedings of the IEEE Conference on Computer Vision and
  Pattern Recognition}, 2019.

\bibitem{Lip2Wav}
K Prajwal, Rudrabha Mukhopadhyay, Vinay Namboodiri, and C Jawahar.
\newblock A lip sync expert is all you need for speech to lip generation in the
  wild, 08 2020.

\bibitem{vae}
Yunchen Pu, Zhe Gan, Ricardo Henao, Xin Yuan, Chunyuan Li, Andrew Stevens, and
  Lawrence Carin.
\newblock Variational autoencoder for deep learning of images, labels and
  captions.
\newblock 09 2016.

\bibitem{pyvoc}
https://github.com/JeremyCCHsu/Python-Wrapper-for-World-Vocoder PyWorldVocoder.

\bibitem{stochastic}
Vinay P Namboodiri Rajesh M~Hegde Ravindra~Yadav, Ashish~Sardana.
\newblock Stochastic talking face generation using latent distribution
  matching.
\newblock 02 2015.

\bibitem{Unet}
Olaf Ronneberger, Philipp Fischer, and Thomas Brox.
\newblock U-net: Convolutional networks for biomedical image segmentation.
\newblock volume 9351, pages 234--241, 10 2015.

\bibitem{Temporalgan}
Masaki Saito, Eiichi Matsumoto, and Shunta Saito.
\newblock Temporal generative adversarial nets with singular value clipping.
\newblock 10 2017.

\bibitem{FOMM}
Aliaksandr Siarohin, Stéphane Lathuilière, Sergey Tulyakov, Elisa Ricci, and
  Nicu Sebe.
\newblock First order motion model for image animation.
\newblock 2020.

\bibitem{SimonCox}
A. Simons and Stephen Cox.
\newblock Generation of mouthshapes for a synthetic talking head.
\newblock {\em Proceedings of the Institute of Acoustics, Autumn Meeting}, 01
  1990.

\bibitem{inproceedings1}
Andreea Stef, Kaveen Perera, Hubert Shum, and Edmond Ho.
\newblock Synthesizing expressive facial and speech animation by text-to-ipa
  translation with emotion control.
\newblock pages 1--8, 12 2018.

\bibitem{hourglass}
Gary Storey, Ahmed Bouridane, Richard Jiang, and Chang-Tsun Li.
\newblock {\em Atypical Facial Landmark Localisation with Stacked Hourglass
  Networks: A Study on 3D Facial Modelling for Medical Diagnosis}, pages
  37--49.
\newblock 01 2020.

\bibitem{article12}
Supasorn Suwajanakorn, Steven Seitz, and Ira Kemelmacher.
\newblock Synthesizing obama: learning lip sync from audio.
\newblock {\em ACM Transactions on Graphics}, 36:1--13, 07 2017.

\bibitem{inproceedings3}
Sarah Taylor, Moshe Mahler, Barry-John Theobald, and Iain Matthews.
\newblock Dynamic units of visual speech.
\newblock pages 275--284, 07 2012.

\bibitem{neuralhead}
Justus Thies, Mohamed Elgharib, Ayush Tewari, Christian Theobalt, and Matthias
  Nießner.
\newblock Neural voice puppetry: Audio-driven facial reenactment, 12 2019.

\bibitem{Tulyakov:2018:MoCoGAN}
Sergey Tulyakov, Ming-Yu Liu, Xiaodong Yang, and Jan Kautz.
\newblock {MoCoGAN}: Decomposing motion and content for video generation.
\newblock In {\em IEEE Conference on Computer Vision and Pattern Recognition
  (CVPR)}, pages 1526--1535, 2018.

\bibitem{isnorm}
Dmitry Ulyanov, Andrea Vedaldi, and Victor Lempitsky.
\newblock Instance normalization: The missing ingredient for fast stylization.
\newblock 07 2016.

\bibitem{SceneDynamics}
Carl Vondrick, Hamed Pirsiavash, and Antonio Torralba.
\newblock Generating videos with scene dynamics.
\newblock 09 2016.

\bibitem{Alpher05}
Konstantinos Vougioukas, Stavros Petridi, and Maja Pantic.
\newblock End-to-end speech-driven facial animation with temporal gans.
\newblock {\em Journal of Foo}, 14(1):234--778, 2004.

\bibitem{eca}
Qilong Wang, Banggu Wu, Pengfei Zhu, Peihua Li, Wangmeng Zuo, and Qinghua Hu.
\newblock Eca-net: Efficient channel attention for deep convolutional neural
  networks, 10 2019.

\bibitem{wang2018pix2pixHD}
Ting-Chun Wang, Ming-Yu Liu, Jun-Yan Zhu, Andrew Tao, Jan Kautz, and Bryan
  Catanzaro.
\newblock High-resolution image synthesis and semantic manipulation with
  conditional gans.
\newblock In {\em Proceedings of the IEEE Conference on Computer Vision and
  Pattern Recognition}, 2018.

\bibitem{Wiles18}
O. Wiles, A.S. Koepke, and A. Zisserman.
\newblock X2face: A network for controlling face generation by using images,
  audio, and pose codes.
\newblock In {\em European Conference on Computer Vision}, 2018.

\bibitem{Yi2020AudiodrivenTF}
R. Yi, Zipeng Ye, J. Zhang, H. Bao, and Yongjin Liu.
\newblock Audio-driven talking face video generation with learning-based
  personalized head pose.
\newblock {\em arXiv: Computer Vision and Pattern Recognition}, 2020.

\bibitem{cam}
B. Zhou, A. Khosla, Lapedriza. A., A. Oliva, and A. Torralba.
\newblock {Learning Deep Features for Discriminative Localization.}
\newblock {\em CVPR}, 2016.

\bibitem{Zhou2019TalkingFG}
Hang Zhou, Y. Liu, Z. Liu, Ping Luo, and X. Wang.
\newblock Talking face generation by adversarially disentangled audio-visual
  representation.
\newblock In {\em AAAI}, 2019.

\bibitem{Zhu2020ArbitraryTF}
Hao Zhu, Huaibo Huang, Y. Li, A. Zheng, and R. He.
\newblock Arbitrary talking face generation via attentional audio-visual
  coherence learning.
\newblock {\em arXiv: Computer Vision and Pattern Recognition}, 2020.

\end{thebibliography}


\begin{thebibliography}{10}\itemsep=-1pt

\bibitem{Alpher03}
FirstName Alpher and FirstName Fotheringham-Smythe.
\newblock Frobnication revisited.
\newblock {\em Journal of Foo}, 13(1):234--778, 2003.

\bibitem{assael2016lipnet}
Yannis~M Assael, Brendan Shillingford, Shimon Whiteson, and Nando de Freitas.
\newblock Lipnet: End-to-end sentence-level lipreading.
\newblock {\em GPU Technology Conference}, 2017.

\bibitem{crema}
Houwei Cao, David Cooper, Michael Keutmann, Ruben Gur, Ani Nenkova, and Ragini
  Verma.
\newblock Crema-d: Crowd-sourced emotional multimodal actors dataset.
\newblock {\em IEEE transactions on affective computing}, 5:377--390, 10 2014.

\bibitem{cao2018openpose}
Zhe Cao, Gines Hidalgo, Tomas Simon, Shih-En Wei, and Yaser Sheikh.
\newblock Open{P}ose: realtime multi-person 2{D} pose estimation using {P}art
  {A}ffinity {F}ields.
\newblock In {\em arXiv preprint arXiv:1812.08008}, 2018.

\bibitem{lmd}
Lele Chen, Zhiheng Li, Ross~K. Maddox, Zhiyao Duan, and Chenliang Xu.
\newblock Lip movements generation at a glance.
\newblock 2018.

\bibitem{Chung17b}
Joon~Son Chung, Amir Jamaludin, and Andrew Zisserman.
\newblock You said that?
\newblock In {\em British Machine Vision Conference}, 2017.

\bibitem{voxceleb}
J.~S. Chung, A. Nagrani, and A. Zisserman.
\newblock Voxceleb2: Deep speaker recognition.
\newblock In {\em INTERSPEECH}, 2018.

\bibitem{facek}
https://github.com/raymon-tian/hourglass-facekeypoints-detection
  facial~keypoint detection.

\bibitem{farneback}
Gunnar Farnebäck.
\newblock Two-frame motion estimation based on polynomial expansion.
\newblock volume 2749, pages 363--370, 06 2003.

\bibitem{resnet}
Kaiming He, Xiangyu Zhang, Shaoqing Ren, and Jian Sun.
\newblock Deep residual learning for image recognition.
\newblock pages 770--778, 06 2016.

\bibitem{patchgan}
Phillip Isola, Jun-Yan Zhu, Tinghui Zhou, and Alexei Efros.
\newblock Image-to-image translation with conditional adversarial networks.
\newblock pages 5967--5976, 07 2017.

\bibitem{PerceptualLoss}
Alexandre~Alahi Justin~Johnson and Li Fei-Fei.
\newblock Perceptual losses for real-time style transfer and super-resolution.
\newblock 2016.

\bibitem{dlib}
D.E. King.
\newblock Dlib-ml: A machine learning toolkit. journal of machine learning
  research.
\newblock 2009.

\bibitem{Kumar_2020_CVPR_Workshops}
Neeraj Kumar, Srishti Goel, Ankur Narang, and Mujtaba Hasan.
\newblock Robust one shot audio to video generation.
\newblock In {\em Proceedings of the IEEE/CVF Conference on Computer Vision and
  Pattern Recognition (CVPR) Workshops}, June 2020.

\bibitem{Authors14}
FirstName LastName.
\newblock The frobnicatable foo filter, 2014.
\newblock Face and Gesture submission ID 324. Supplied as additional material
  {\tt fg324.pdf}.

\bibitem{RLoss}
Yanchun Li, Nanfeng Xiao, and Wanli Ouyang.
\newblock Improved generative adversarial networks with reconstruction loss.
\newblock {\em Neurocomputing}, 323, 10 2018.

\bibitem{acd2}
Yu~Tian-Mubbasir~Kapadia Long~Zhao, Xi~Peng and Dimitris Metaxas1.
\newblock Learning to forecast and refine residual motion for image-to-video
  generation, 2018.

\bibitem{VGGFace}
Wang Mei and Weihong Deng.
\newblock Deep face recognition: A survey.
\newblock 04 2018.

\bibitem{spectralnorm}
Takeru Miyato, Toshiki Kataoka, Masanori Koyama, and Yuichi Yoshida.
\newblock Spectral normalization for generative adversarial networks.
\newblock 02 2018.

\bibitem{gridlombard}
Ricard Marxer-Jon~Barker Najwa~Alghamdi, Steve~Maddock and Guy~J. Brown.
\newblock A corpus of audio-visual lombard speech with frontal and profile
  view, the journal of the acoustical society of america 143, el523 (2018);
  https://doi.org/10.1121/1.5042758, 2018.

\bibitem{VGG19}
Karen Simonyan and Andrew Zisserman.
\newblock Very deep convolutional networks for large-scale image recognition.
\newblock {\em arXiv 1409.1556}, 09 2014.

\bibitem{Tulyakov:2018:MoCoGAN}
Sergey Tulyakov, Ming-Yu Liu, Xiaodong Yang, and Jan Kautz.
\newblock {MoCoGAN}: Decomposing motion and content for video generation.
\newblock In {\em IEEE Conference on Computer Vision and Pattern Recognition
  (CVPR)}, pages 1526--1535, 2018.

\bibitem{Alpher05}
Konstantinos Vougioukas, Stavros Petridi, and Maja Pantic.
\newblock End-to-end speech-driven facial animation with temporal gans.
\newblock {\em Journal of Foo}, 14(1):234--778, 2004.

\bibitem{pix2pixhd}
Ting-Chun Wang, Ming-Yu Liu, Jun-Yan Zhu, Andrew Tao, Jan Kautz, and Bryan
  Catanzaro.
\newblock High-resolution image synthesis and semantic manipulation with
  conditional gans.
\newblock In {\em Proceedings of the IEEE Conference on Computer Vision and
  Pattern Recognition}, 2018.

\bibitem{wang2018pix2pixHD}
Ting-Chun Wang, Ming-Yu Liu, Jun-Yan Zhu, Andrew Tao, Jan Kautz, and Bryan
  Catanzaro.
\newblock High-resolution image synthesis and semantic manipulation with
  conditional gans.
\newblock In {\em Proceedings of the IEEE Conference on Computer Vision and
  Pattern Recognition}, 2018.

\end{thebibliography}

\end{document}


\title{Supplementary Material for Multi Modal Adaptive Normalization for Audio to Video Generation}

\maketitle

\section{Architectural details}
\subsection{Generator} 
\begin{figure}[h!]
\centering
  \includegraphics[width=0.6\linewidth]{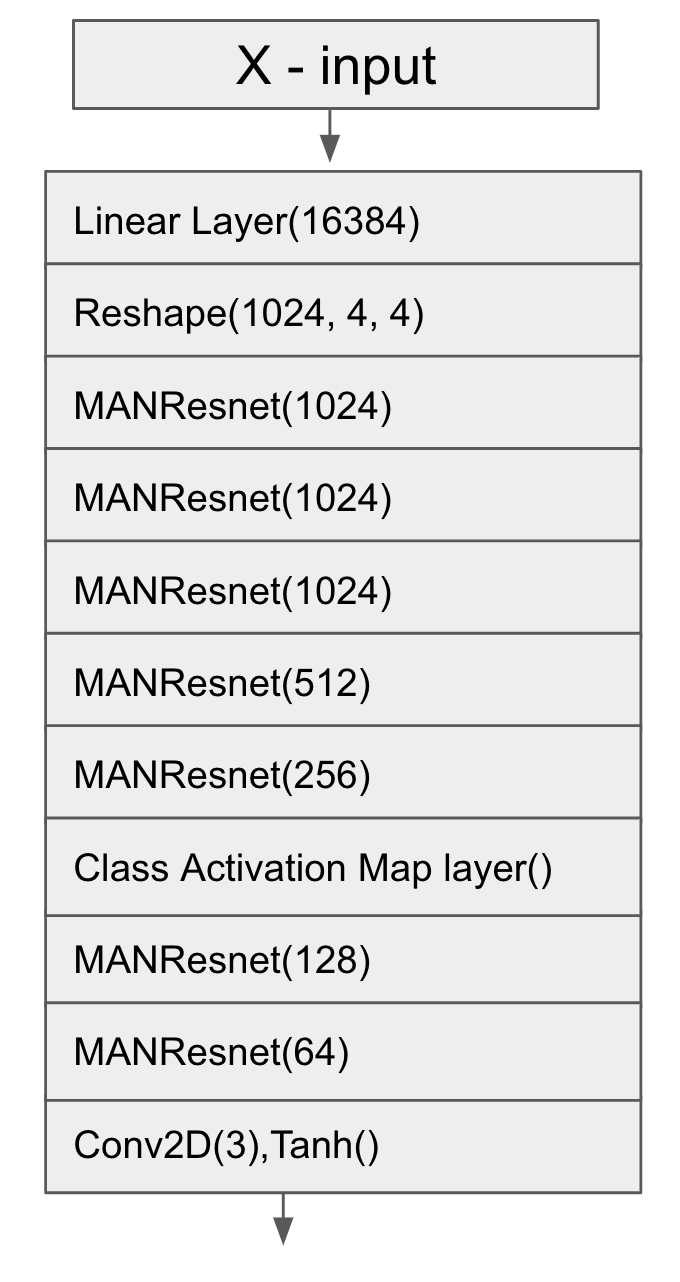}
  \caption{Generator Architecture}
  \label{fig:gen}
\end{figure}

\begin{figure}[h!]
  \includegraphics[width=\linewidth]{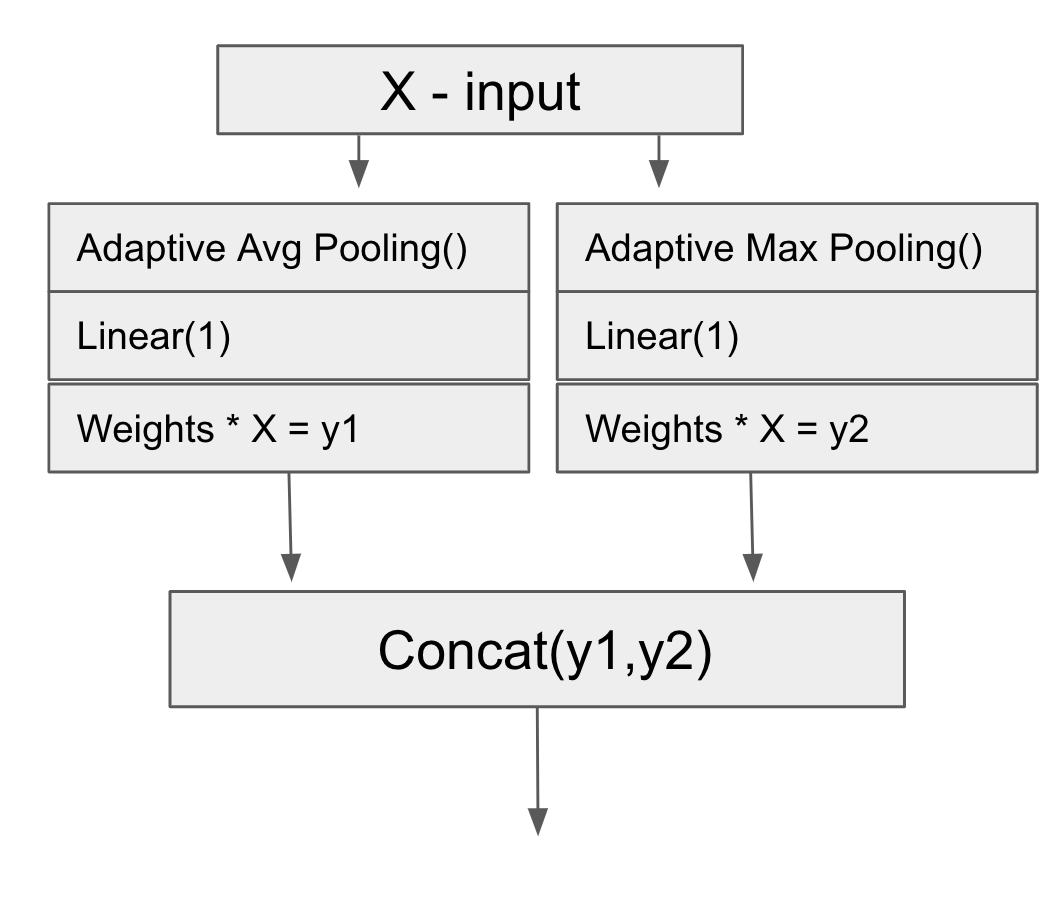}
  \caption{Class Activation Map based layer}
  \label{fig:caml}
\end{figure}

We have used the generator which is based on multi modal adaptive normalization based architecture. We use the synchronized version of the BatchNorm. We apply the Spectral Norm\cite{spectralnorm} to all the convolution layers in the generator in Figure ~\ref{fig:gen}. X is the 13 dimensional mel audio features which goes into the initial layer of generator. The class Activation map based layers uses global average pooling and global max pooling to focus on global and local features needed for feature generation in Figure ~\ref{fig:caml}. The X input in class activation based layer is the previous feature map coming from multi modal adaptive normalization based residual block.

\begin{figure}[h!]
  \includegraphics[width=\linewidth]{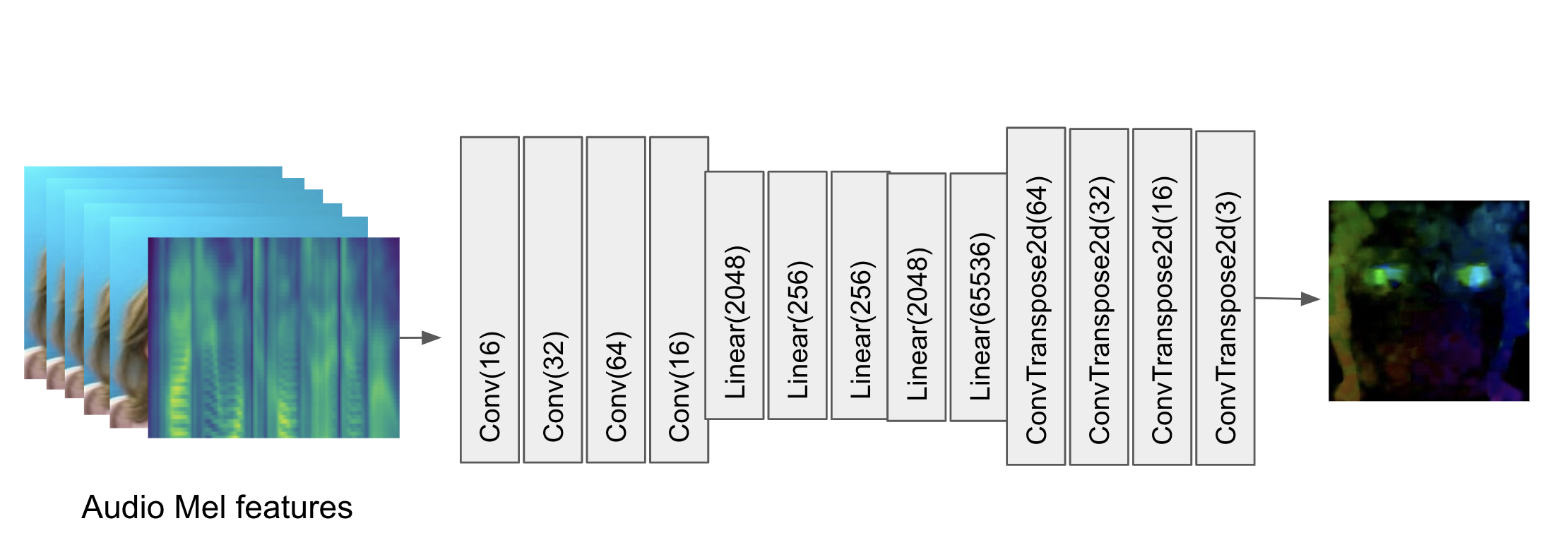}
  \caption{Optical flow predictor Architecture}
  \label{fig:optigen}
\end{figure}

\paragraph{Optical flow predictor}
We have used the Figure ~\ref{fig:optigen} architecture which gives previous 5 frames and 256 dimensional mel audio features to generate the next optical flow used in multi modal adaptive normalization.

\begin{figure}[h!]
  \includegraphics[width=\linewidth]{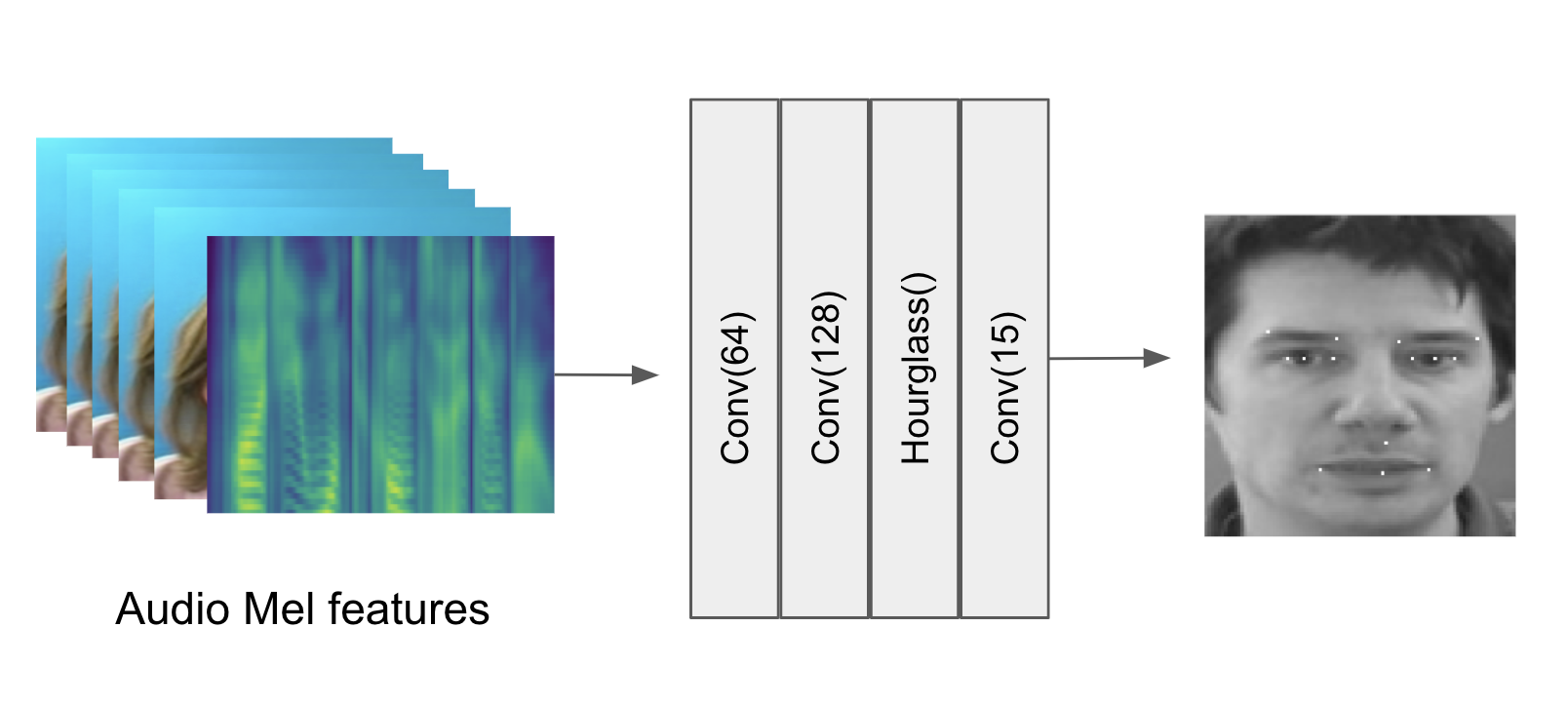}
  \caption{Keypoint heatmap  predicted Architecture}
  \label{fig:keygen}
\end{figure}

\paragraph{Keypoint heatmap  predictor}
We have used the Figure ~\ref{fig:keygen} architecture which gives previous 5 frames and 256 dimensional mel audio features to generate the 15 channel keypoint heatmap used in multi modal adaptive normalization. This is optimization  with mean square error loss with pretrained facial keypoint detection model\cite{facek}.

\begin{figure}[h!]
  \includegraphics[width=\linewidth]{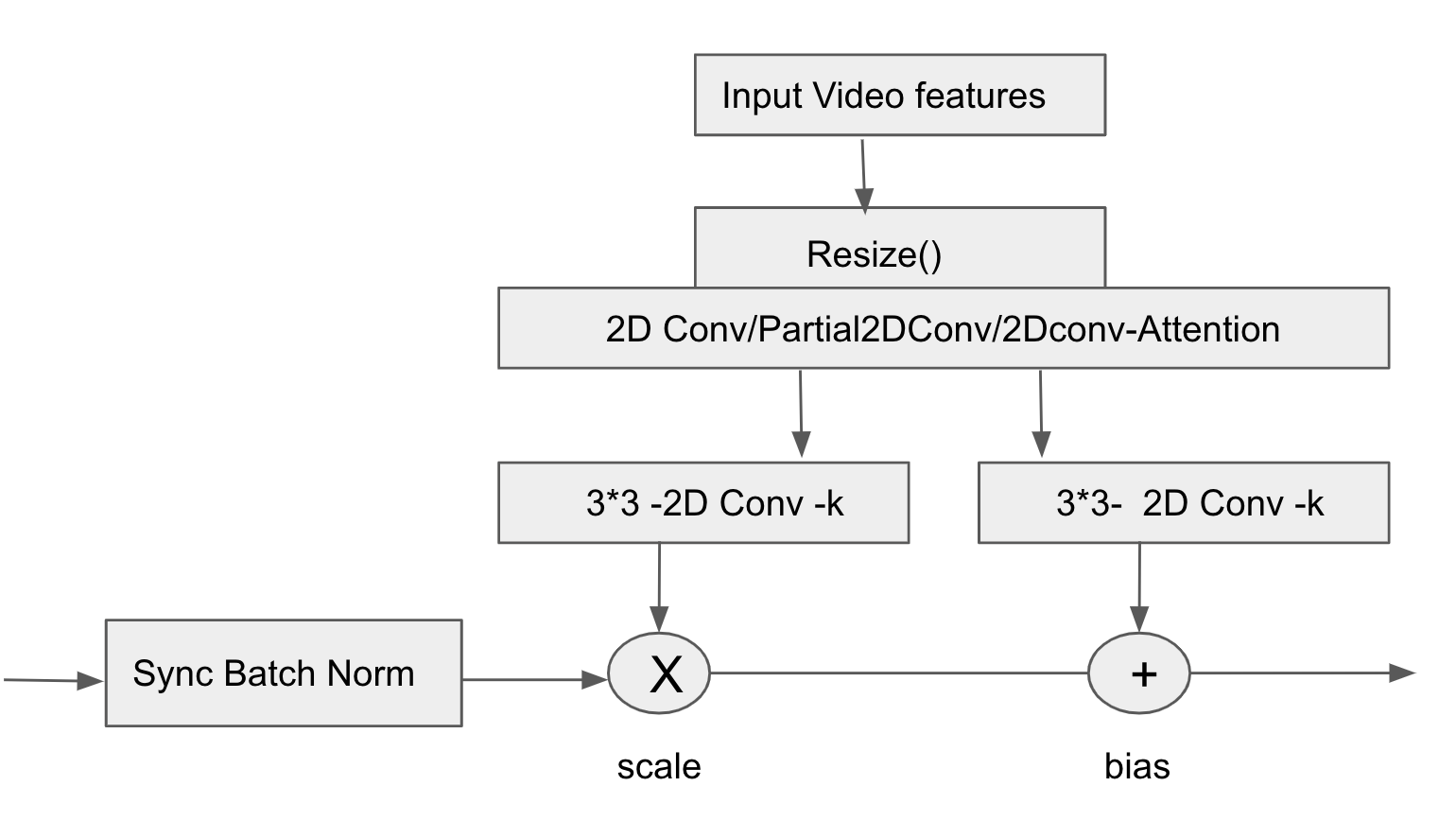}
  \caption{Architecture to calculate the affine parameters from video features in Multi modal adaptive normalization }
  \label{fig:manvid}
\end{figure}

\begin{figure}[h!]
  \includegraphics[width=\linewidth]{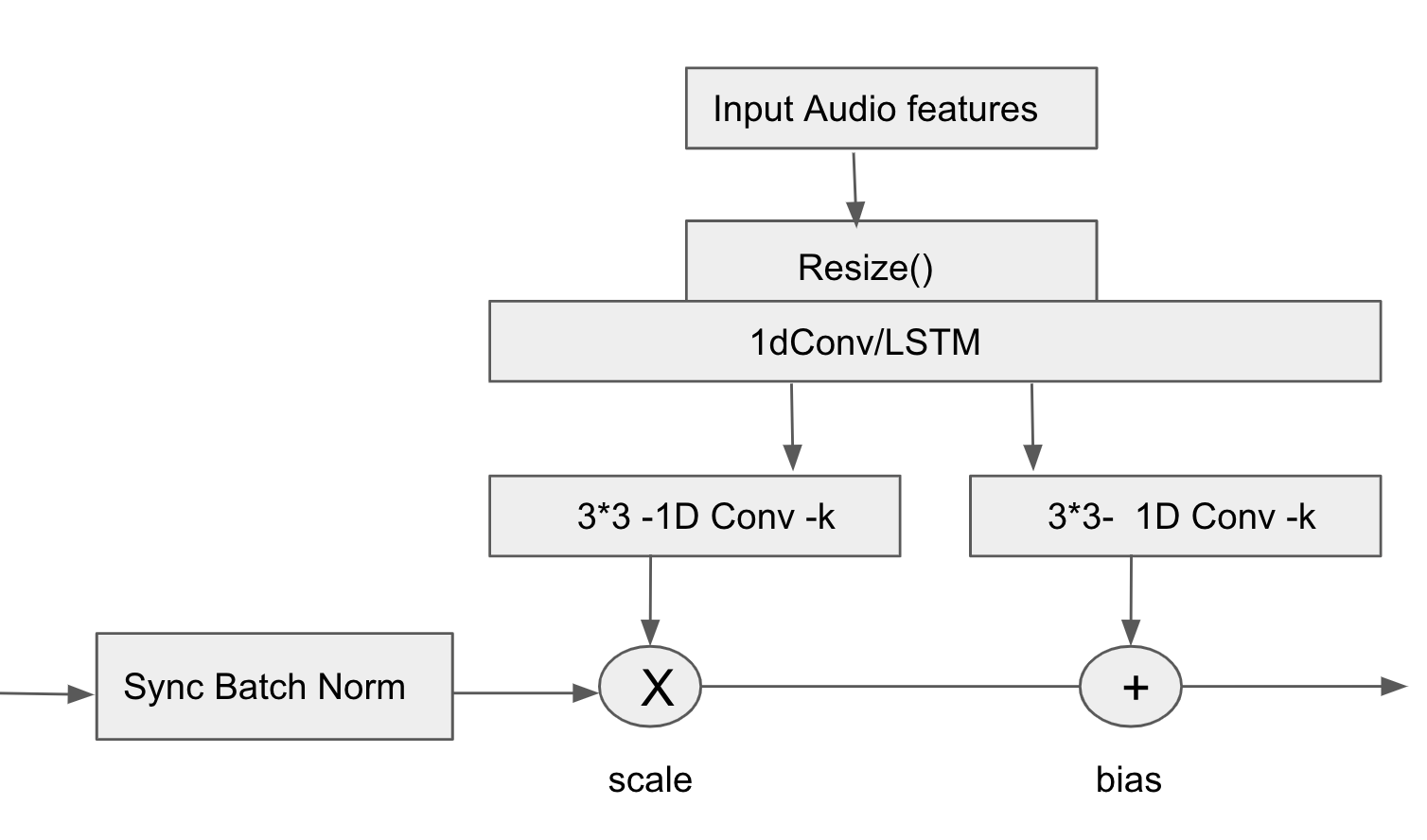}
  \caption{Architecture to calculate the affine parameters from audio features in Multi modal adaptive normalization }
  \label{fig:manaud}
\end{figure}

\paragraph{Multi modal adaptive normalization}
This novel architecture takes the various audio and video features namely person's image, predicted optical flow/predicted keypoint heatmap, mel spectrogram features, pitch and energy. The video features i.e. person's image, predicted optical flow/predicted keypoint heatmap goes into the normalization to calculate the respective affine parameters  as shown in Figure ~\ref{fig:manvid}. The audio features i.e. mel spectrogram features, pitch and energy, calculate the affine parameters through the architecture as shown in Figure ~\ref{fig:manaud}. The amount of information used by model from this normalization is controlled by  various $\rho$'s which is fed into the softmax function as shown in (Equation~\eqref{rhosum1})

 \begin{equation}
 \label{rhosum1}
   \rho\textsubscript{1}+\rho\textsubscript{2}+\rho\textsubscript{3}+\rho\textsubscript{4}+\rho\textsubscript{5} = 1
\end{equation}

\begin{figure}[h!]
  \includegraphics[width=\linewidth]{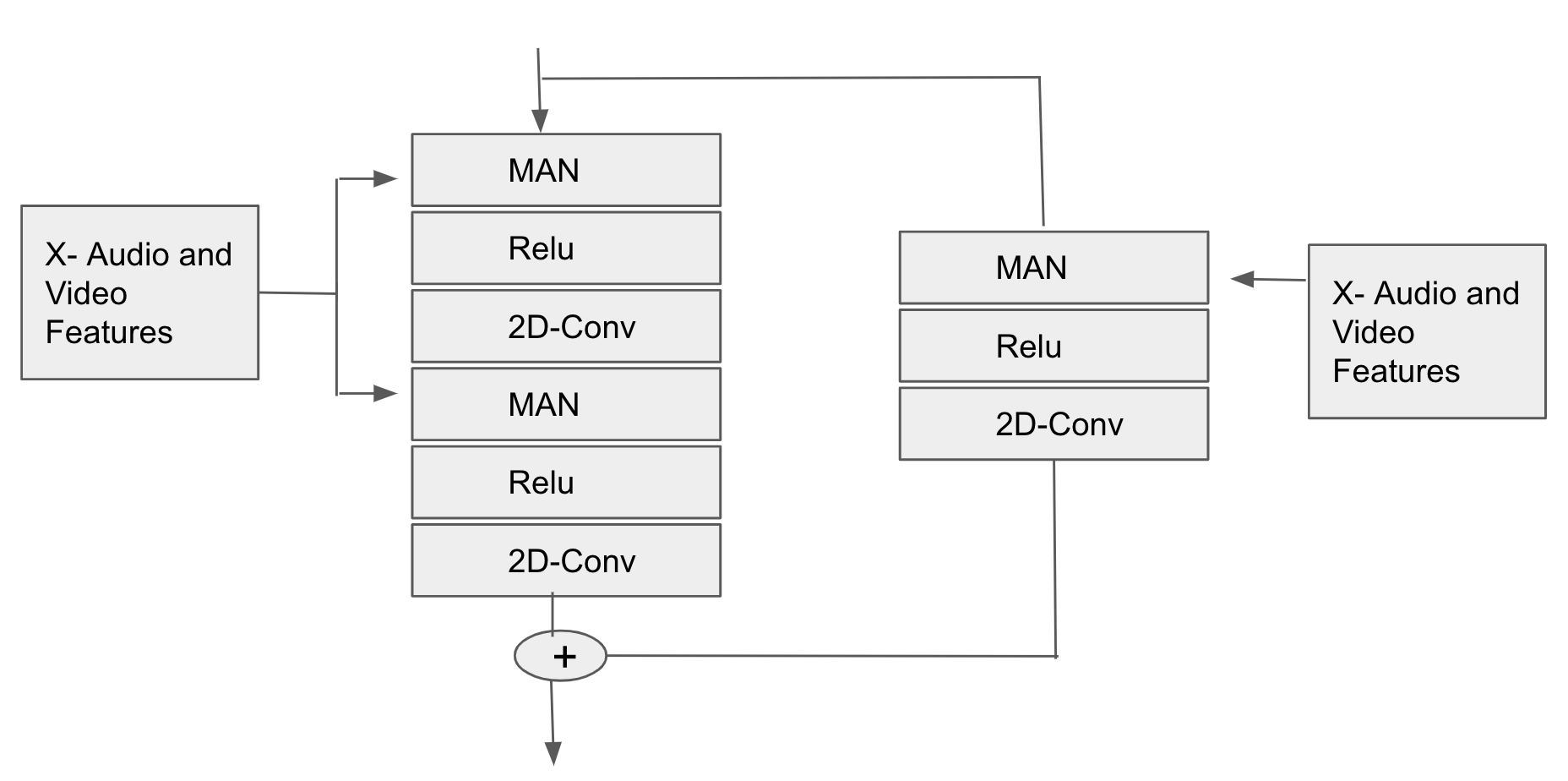}
  \caption{Multi modal adaptive normalization resnet  Architecture}
  \label{fig:manresblk}
\end{figure}

\paragraph{MAN Resnet}
Figure ~\ref{fig:manresblk} shows the resnet\cite{resnet} architecture around multi modal adaptive normalization(MAN) where X represents the audio and video features namely person's image, predicted optical flow/predicted keypoint heatmap, mel spectrogram features, pitch and energy which goes into the multi modal adaptive normalization network.

\subsection{Discriminator}

\begin{figure}[h!]
  \includegraphics[width=\linewidth]{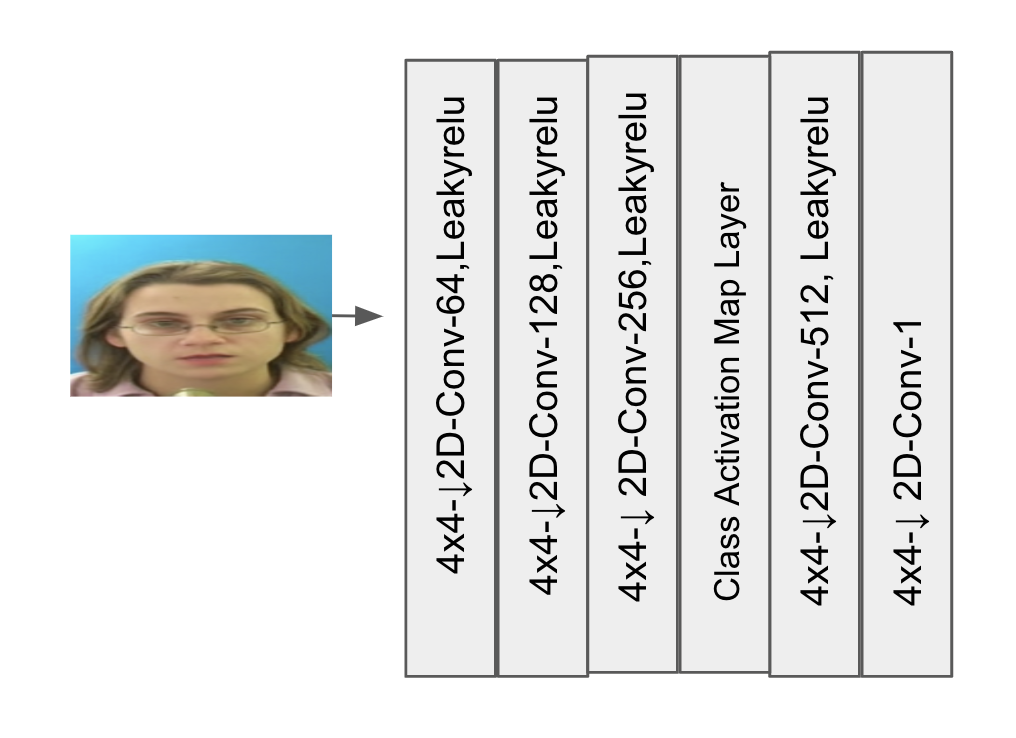}]
  \caption{Discriminator Architecture}
  \label{fig:patchgan}
\end{figure}

This multi-scale frame discriminator is based on Pix2PixHD\cite{pix2pixhd}. We have used spectral normalization\cite{spectralnorm} instead of instance normalization. It is based on the PatchGAN\cite{patchgan}. Hence, the last layer of the discriminator is a convolution layer as shown in Figure ~\ref{fig:patchgan}.

\section{ Experimentation Details}

 \subsection {Losses}
The proposed method is trained with different losses to generate realistic videos as explained below.

\subsubsection{Adversarial Loss}
Adversarial Loss is used to train the model to handle adversarial attacks and ensure generation of high-quality images for the video. The loss is defined as:

\begin{equation}
    L\textsubscript{GAN}(G,D) = E\textsubscript{x$\sim$P\textsubscript{d}}[\log(D(x))] + E\textsubscript{z$\sim$P\textsubscript{z}}[\log(D(1-G(z)))] 
\end{equation}

 where G tries to minimize this objective against an adversarial D that tries to maximize.

\subsubsection{Reconstruction loss}
Reconstruction loss~\cite{RLoss} is used on the lower half of the image to improve the reconstruction in mouth area. L1 loss is used for this purpose as described below:

\begin{equation}
    L\textsubscript{RL} = \sum_{n\epsilon [0,W]*[H/2,H]}^{}(R\textsubscript{n} - G\textsubscript{n})
\end{equation}

where, R\textsubscript{n} and G\textsubscript{n} are the real and generated frames respectively.

\subsubsection{Feature Loss}

Feature-matching Loss~\cite{wang2018pix2pixHD} ensures generation of natural-looking  high-quality frames. We take the L1 loss of between generated images and real images for different scale discriminators and then sum it all. We extract features from multiple layers of the discriminator and learn to match these intermediate representations from the real and the synthesized image. This helps in stabilizing the training of the generator. The feature matching loss,   L\textsubscript{FM}(G,D\textsubscript{k}) is given by:

\begin{equation}
    L\textsubscript{FM}(G,D\textsubscript{k}) = E\textsubscript{(x,z)} \sum_ {n=1}^{T}[\frac{1}{N\textsubscript{i}}||D\textsubscript{k}^{(i)}(x)-D\textsubscript{k}^{(i)}(G(z))||\textsubscript{1}]
\end{equation}

where, T is the total number of layers and N\textsubscript{i} denotes the
number of elements in each layer.

\subsubsection{Perceptual Loss}
The perceptual similarity metric is calculated between the generated frame and the real frame. This is done by using features of a VGG19~\cite{VGG19} model trained for ILSVRC classification and VGGFace~\cite{VGGFace} dataset.The perceptual loss~\cite{PerceptualLoss},(L\textsubscript{PL}) is defined as:

\begin{equation}
    L\textsubscript{PL} = \lambda\sum_{n=1}^{N}[\frac{1}{M\textsubscript{i}}||F^{(i)}(x)-F^{(i)}(G(z))||\textsubscript{1}]
\end{equation}
where, $\lambda$ is the weight for perceptual loss and $F^{(i)}$ is the ith layer of VGG19 network with M\textsubscript{i} elements of VGG layer.

\subsubsection{Class Activation Loss} We have used the class activation based adversarial loss in generator and discriminator which help the model to learn local and global facial features and helps in cheeck movement, blinks as well as image reconstruction.

 \begin{equation}
 \label{camloss}
    L\textsubscript{cam} = E\textsubscript{y$\sim$P\textsubscript{t}}[\log(n\textsubscript{D\textsubscript{t}}(y))] + E\textsubscript{x$\sim$P\textsubscript{s}}[\log(D(1-n\textsubscript{D\textsubscript{t}}(G(x))))] 
\end{equation}

where,$n\textsubscript{D\textsubscript{t}}$ is the class activation based logits from real and fake image .

\subsubsection{Mean Square loss}
We have optimized the keypoint heatmap predictor and optical flow predictor using mean square loss between generated keypoint heatmap and pretrained model\cite{facek} and generated optical flow and ground truth farneback\cite{farneback} optical flow output.

\subsection{Evaluation Metric}
\paragraph{1. PSNR- Peak Signal to Noise Ratio:} It computes the peak signal to noise ratio between two images. The higher the PSNR the better the quality of the reconstructed image.

\paragraph{2. SSIM- Structural Similarity Index:} It is a perceptual metric that quantifies image quality degradation.\ The larger the value the better the quality of the reconstructed image.

\paragraph{3. CPBD- Cumulative Probability Blur Detection:} It is a perceptual based no-reference objective image sharpness metric based on the cumulative probability of blur detection developed at the Image.

\paragraph{4. WER- Word error rate:} It is a metric to evaluate the performance of speech recognition in a given video. We have used LipNet architecture~\cite{assael2016lipnet} which is pre-trained on the GRID dataset for evaluating the WER. On the GRID dataset, Lipnet achieves 95.2 percent accuracy which surpasses the experienced human lipreaders.

\paragraph{5. ACD- Average Content Distance(~\cite{Tulyakov:2018:MoCoGAN}):} It is used for the identification of speakers from the generated frames using OpenPose~\cite{cao2018openpose}. We have calculated the Cosine distance and Euclidean distance of representation of the generated image and the actual image from Openpose. The distance threshold for the OpenPose model should be 0.02 for Cosine distance and 0.20 for Euclidean distance ~\cite{acd2}. The lesser the distances the more similar the generated and actual images.

\paragraph{6. LMD - Landmark Distance(~\cite{lmd}):} To ensure realistic and accurate lip movement, ensuring good performance on speech recognition we use this metric. We calculate the landmark points~\cite{dlib} on both real and generated images at the scale of 256*256 and use the lip region points i.e., points 49-68 and call then as LR and LF respectively.T is the number of frames. Then, we calculate the euclidean distance between each corresponding pairs of landmarks on LR and LF. The LMD is defined as: 

\begin{equation}
    LMD = \frac{1}{T} X \frac{1}{P} \sum_{t=1}^{T} \sum_{p=1}^{P} ||LR\textsubscript{t,p} - LF\textsubscript{t,p}||
\end{equation} 

\paragraph{7. Blinks/sec:} To capture the blinks in the video, we are calculating the blinks/sec so that we can better understand the quality of animated videos. We have used SVM and eye landmarks along with Eye aspect ratio used in  Real-Time Eye Blink Detection using Facial Landmarks ~\cite{Authors14} to detect the blinks in a video.

\subsection{Datasets}

\paragraph{GRID}
GRID~\cite{Alpher03} is a large multi-talker audiovisual sentence corpus to support joint computational-behavioral studies in speech perception. In brief, the corpus consists of high-quality audio and video (facial) recordings of 1000 sentences spoken by each of 34 talkers (18 male, 16 female). Sentences are of the form "put red at G9 now".

\paragraph{Lombard Grid}
Lombard Grid~\cite{gridlombard} is a bi-view audiovisual Lombard speech corpus that can be used to support joint computational-behavioral studies in speech perception. The corpus includes 54 talkers, with 100 utterances per talker (50 Lombard and 50 plain utterances). This dataset follows the same sentence format as the audiovisual Grid corpus, and can thus be considered as an extension of that corpus.

\paragraph{CREMA-D}
CREMA-D~\cite{crema} is a data set of 7,442 original clips from 91 actors. These clips were from 48 male and 43 female actors between the ages of 20 and 74 coming from a variety of races and ethnicities (African American, Asian, Caucasian, Hispanic, and Unspecified). Actors spoke from a selection of 12 sentences. The sentences were presented using one of six different emotions (Anger, Disgust, Fear, Happy, Neutral, and Sad) and four different emotion levels (Low, Medium, High, and Unspecified).

\paragraph{VoxCeleb2}
VoxCeleb2\cite{voxceleb} is a very large-scale audio-visual speaker recognition dataset collected from open-source media. Voxceleb2 contains over 1 million utterances for over 6,000 celebrities, extracted from videos uploaded to YouTube. The dataset is fairly gender balanced, with 61 \% of the speakers male. The speakers span a wide range of different ethnicities, accents, professions and ages.

\subsection{Training Details}

We have trained the model with 5000 videos in training and used 1000 videos in test for all the 4 datasets namely Grid, Grid LOMBARD, CREMA-D and VoxCeleb2. We train the model for 10 epochs for Grid and Grid LOMBARD, 20 epochs for CREMA-D and 50 epochs for VoxCeleb2 dataset. The frames are resized into 256X256 size and audio features are processed at 16khz.

\section{Psychophysical assessment}

\begin{figure}[h!]
  \includegraphics[width=\linewidth]{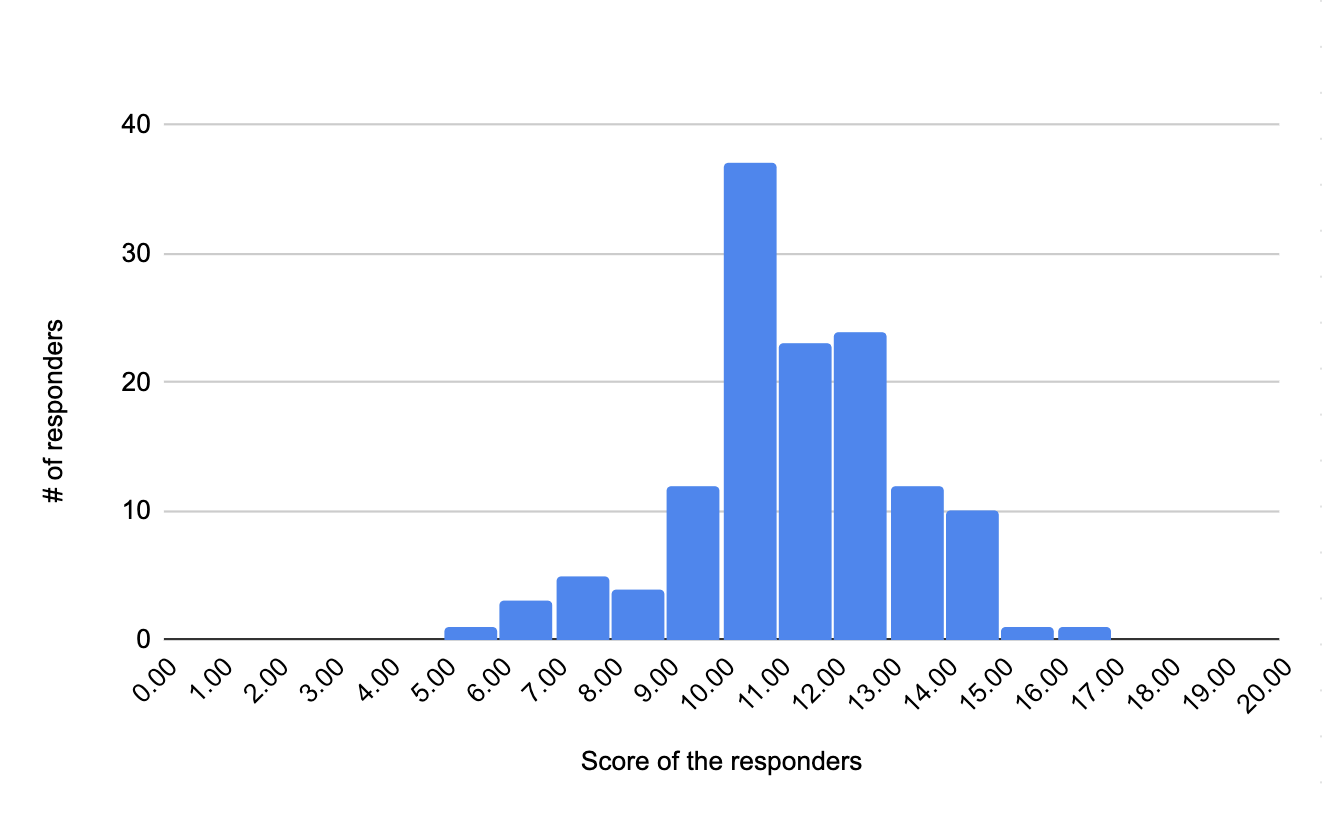}
  \caption{Distribution of user scores for the online Turing test}
  \label{fig:turing}
\end{figure}

To test the naturalism of the generated videos we conduct an online Turing test on GRID dataset \footnote{\url{https://forms.gle/DM1DRcTToQFvUpTa7}}. Each test consists of 20 questions with 10 fake and 10 real videos. The user is asked to label a video real or fake based on the aesthetics and naturalism of the video. Approximately 300 user data is collected and their score of the ability to spot fake video is displayed in Figure ~\ref{fig:turing}.

Results are visually rated (on a scale of 5) individually by 25 persons, on three aspects,
 lip synchronization, eye blinks and eyebrow raises and quality of video on Grid dataset.
 The subjects were shown anonymous videos at the same time for the different audio clips for side-by-side comparison. Table ~\ref{tab:table1} clearly shows that MAN based proposed architecture performs significantly better in quality and lip synchronization which is of prime importance in videos. 


\begin{table}[h!]
  \begin{center}

    \begin{tabular}{c|c|c|c} 
      \textbf{Method} & \textbf{Lip-Sync} & \textbf{Eye-blink} & \textbf{Quality}\\
      \hline
       MAN & 91.8 & \textbf{90.5} & \textbf{79.6}  \\
       OneShotA2V\cite{Kumar_2020_CVPR_Workshops}& 90.8 & 88.5 & 76.2  \\
       RSDGAN\cite{Alpher05}  & \textbf{92.8} & 90.2 & 74.3\\
       Speech2Vid\cite{Chung17b}  & 90.7 & 87.7 & 72.2 \\
    \end{tabular}
    \vspace {0.25\baselineskip}
    \caption{Psychophysical Evaluation (in percentages) based on users rating on Grid daatset}
    \label{tab:table1}
  \end{center}
\end{table}


s